\pdfoutput=1

\documentclass[11pt]{article}
\PassOptionsToPackage{prologue,dvipsnames}{xcolor}
\usepackage[final]{acl}

\usepackage{times}
\usepackage{latexsym}

\usepackage[T1]{fontenc}

\usepackage[utf8]{inputenc}

\usepackage{microtype}
\usepackage{comment}
\usepackage{inconsolata}

\usepackage{hyperref}
\usepackage{graphicx}
\usepackage{algorithm}
\usepackage{algorithmic}
\usepackage{booktabs}

\usepackage{amssymb}  
\usepackage{xcolor}
\usepackage{colortbl}
\usepackage{adjustbox}

\newcommand{\up}[1]{\textcolor{PineGreen}{\small \ $\uparrow${#1}}}

\newcommand{\down}[1]{\textcolor{PineGreen}{\small \ $\downarrow${#1}}}
\definecolor{darkgreen}{RGB}{0, 100, 0}  
\definecolor{darkred}{RGB}{180, 0, 0}  
\usepackage[most]{tcolorbox}
\usepackage{multirow}
\usepackage{booktabs}

\usepackage{amsmath,amsfonts,bm}

\def\eqref#1{equation~\ref{#1}}

\def\1{\bm{1}}

\DeclareMathAlphabet{\mathsfit}{\encodingdefault}{\sfdefault}{m}{sl}
\SetMathAlphabet{\mathsfit}{bold}{\encodingdefault}{\sfdefault}{bx}{n}

\usepackage{fancyvrb}
\usepackage{commenting}
\usepackage{cite}
\usepackage{subcaption}
\usepackage{longtable}  
\usepackage{amsmath,lipsum}
\usepackage{cuted}
\usepackage{array}  

\title{Understanding the Repeat Curse in Large Language Models \\from a Feature Perspective}

\author{
  \textbf{Junchi Yao\textsuperscript{1,2,3,*}}, 
    \textbf{Shu Yang\textsuperscript{1,2,*}}, \textbf{Jianhua Xu\textsuperscript{1,2,3}}, \\
    \textbf{Lijie Hu\textsuperscript{1,2}}, 
    \textbf{Mengdi Li\textsuperscript{1,2}}, 
    \textbf{Di Wang\textsuperscript{1,2,†}}\\
  \textsuperscript{1}Provable Responsible AI and Data Analytics (PRADA) Lab, \\
  \textsuperscript{2}King Abdullah University of Science and Technology, \\
  \textsuperscript{3}University of Electronic Science and Technology of China\\
  }

\begin{document}
\maketitle

\begin{abstract}
Large language models (LLMs) have made remarkable progress in various domains, yet they often suffer from repetitive text generation, a phenomenon we refer to as the ``Repeat Curse''. While previous studies have proposed decoding strategies to mitigate repetition, the underlying mechanism behind this issue remains insufficiently explored. In this work, we investigate the root causes of repetition in LLMs through the lens of mechanistic interpretability. Inspired by recent advances in Sparse Autoencoders (SAEs), which enable monosemantic feature extraction, we propose a novel approach—``Duplicatus Charm''—to induce and analyze the Repeat Curse. Our method systematically identifies ``Repetition Features'' -the key model activations responsible for generating repetitive outputs. First, we locate the layers most involved in repetition through logit analysis. Next, we extract and stimulate relevant features using SAE-based activation manipulation. To validate our approach, we construct a repetition dataset covering token and paragraph level repetitions and introduce an evaluation pipeline to quantify the influence of identified repetition features. Furthermore, by deactivating these features, we have effectively mitigated the Repeat Curse. The source code of our work is publicly available at: \url{https://github.com/kaustpradalab/repeat-curse-llm}
\end{abstract}
\def\thefootnote{*}\footnotetext{Equal Contribution. The order of these two authors follows alphabetical order of their last names.}
\def\thefootnote{†}\footnotetext{Corresponding Author}
\section{Introduction}
\begin{figure}[htbp]
\centering
\includegraphics[width=0.5\textwidth]{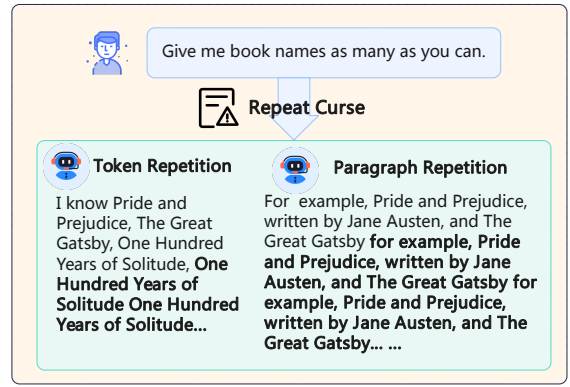}  
\caption{Examples of Repeat Curse: (a) Token Repetition Scenario, (b) Paragraph Repetition Scenario.}
\label{fig:repeatcurse} 
\end{figure}

Large language models (LLMs) have demonstrated remarkable progress across various domains, from machine translation~\citep{xu2024contrastive,wang2023document} and open-ended text generation~\citep{carlsson2024hyperfitting,lee2022factuality,su2023detectllm,su2023fake} to interdisciplinary applications in social science behavior analysis~\citep{yang2025fraud,yao2024fusing,park2023generative,yangmodel} and psychological research~\citep{hu2024psycollm,demszky2023using,yang2024makes}.
Although LLMs have been extensively studied, a critical phenomenon that limits their practical utility is their tendency to generate repetitive content~\citep{fu2021theoretical,xue2024repeat,wang2024mitigating}, which is particularly evident in enumerative tasks, ultimately reducing the performance and diversity of the generated outputs.  We refer to this issue as \textbf{``Repeat Curse''} (see Figure \ref{fig:repeatcurse} for examples).

Previous research has investigated the phenomenon of repetition and has proposed strategies to reduce its occurrence from the perspective of decoding. For example, 
\citet{zhu-etal-2023-penalty} analyzed the self-reinforcement effect in text generation and proposed a repetition penalty mechanism to mitigate its impact. \citet{holtzman2019curious} proposed Nucleus Sampling as a decoding strategy for language models, which can reduce repetition in long texts and improve the generation quality. 
Though these methods can reduce repetition, they may compromise the model's overall performance, and the underlying mechanisms driving repetitive content generation in LLMs remain scarcely studied.

To address the issue, a few works identify the most important component in the network of the Repeat Curse from the mechanistic interpretability view.  
\citet{vaidya-etal-2023-humans} identified specific attention heads and layers that tend to copy the next token by examining the model’s attention maps.
Building upon the layers, \citet{hiraoka2024repetitionneuronslanguagemodels} identified the repetition neurons by analyzing the activation outputs in the feed-forward network of each layer.

Compared to the model neurons, Sparse Autoencoders (SAEs) have been used in LLMs \citep{bricken2023monosemanticity, cunningham2023sparse} to achieve monosemantic units of analysis. SAE maps the complex superposition of polysemantic neurons into monosemantic features. By regularizing activations, it ensures that only a small set of features are activated for each input, making the resulting features human-interpretable \citep{yun2021transformer, rajamanoharan2024improving}. Due to its advances, many recent studies have leveraged SAE to provide the most critical and human-understandable features for different tasks. For example, \citet{le2024learningbiologicallyrelevantfeatures, simon2024interplmdiscoveringinterpretablefeatures}  identified biological-relevant features and further utilized through SAE; \citet{kim2025conceptsteerersleveragingksparse} identified features related to inappropriate content such as nudity and violence. Inspired by these works, we pose the following research question: \textit{Can we leverage SAE to identify the features that cause the Repeat Curse to give a better understanding?}


Unlike the above-mentioned work, we cannot directly identify specific words or phrases that represent repetition. Therefore, to identify these features, we propose the \textbf{``Duplicatus Charm''} \textit{(a magic spell inspired by Harry Potter)} to induce the Repeat Curse.

The main difficulties of our method are locating and identifying the ``target of the spell'', i.e., the most significant features. To address the first problem,  we first analyze the logits to identify the layers that have a significant impact on predicting the next token as a repeat token \citep{nostalgebraist2020logitlens}. Then, we explore the features in those layers by stimulating their activations through SAE \textbf{(\S \ref{par:identification})}. For the second challenge, we design a pipeline to evaluate the effectiveness of the magic spell. First, we construct a repetition dataset containing two scenarios \textbf{(\S \ref{par:dataset})} and then select the appropriate repeat score for the task through the dataset \textbf{(\S \ref{par:metric})}. Leveraging such a metric enables us to pinpoint the features that are most responsible for inducing repetition. We refer to these "targets of the spell" as \textbf{``Repetition Features''}.
Finally, we cast a spell on the repetition features and scored them using the repeat score we identified. From a data perspective, this allows us to demonstrate whether our spell is effective while manually reviewing the texts with higher scores.


We select three language models with different scales: GPT2-small \citep{gpt2}, Gemma-2-2B \citep{gemma_2024}, and Llama-3.1-8B \citep{llama3}. The results show that repetition features are primarily located in all three models' intermediate and final layers, suggesting a consistent pattern across different model architectures and scales.  With the same coefficient, we demonstrate that activating these features increases repetition while other standard features do not. Moreover, deactivating these features could mitigate the Repeat Curse, without exerting any detrimental effects on the model's performance. Leveraging the human-readable nature of these features, we can also summarize the repetition feature's characteristics. 
Overall, our contributions are as follows:


\begin{itemize}
    \item We revisited the phenomenon of the LLM Repeat Curse and uncovered a potential reason why such repetition occurs: the presence of repetition features.
    \item From an interpretability perspective, we proposed a practical and effective pipeline for extracting repetition features.
    \item Our research has been rigorously validated through a series of comprehensive experiments, which confirm the validity and effectiveness of our findings. It deepens our understanding of repetition in LLMs and offers new directions for their optimization.
\end{itemize}
\begin{figure*}[htbp]
\centering
\includegraphics[width=0.96\textwidth]{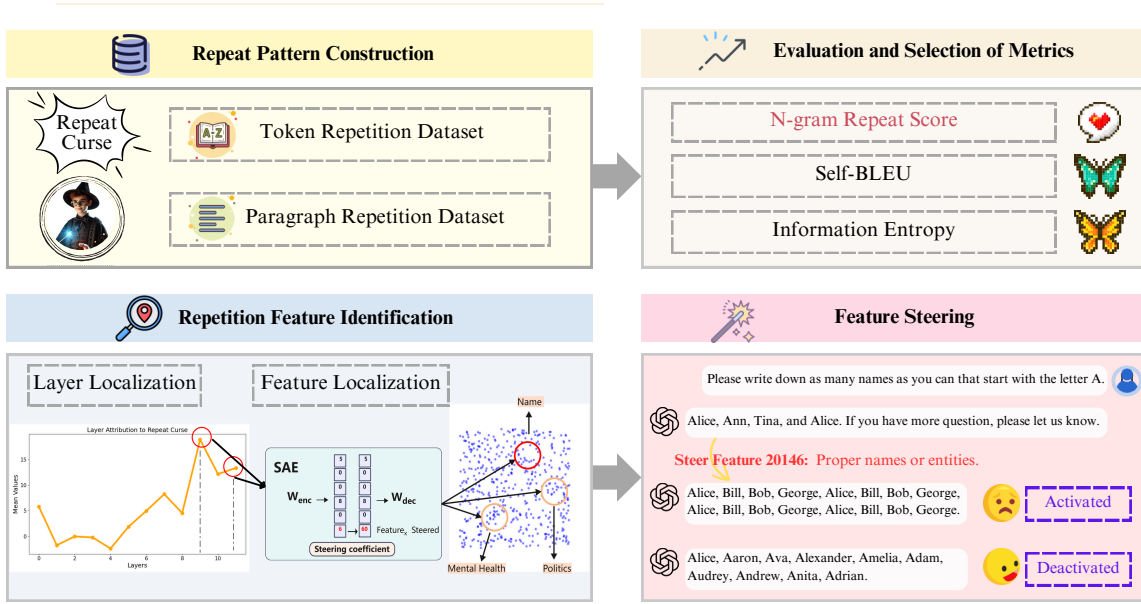}
\vspace{-15pt}
\caption{Illustration of our work (using GPT as an example). \textbf{First line:} The Repeat Curse is categorized into two scenarios: Token and Paragraph, and datasets are created accordingly. These datasets are used to evaluate and select the metrics. \textbf{Second line:} The identification of Repetition Features is divided into two steps: layer localization and feature localization. By identifying the repetition features, we can deactivate them to mitigate the Repeat Curse.}  
\label{fig:method}
\vspace{-7pt}
\end{figure*}
\section{Related Work}

\noindent {\bf Repetation in Language Models.} 
Repetition in language models refers to the phenomenon where the generated text exhibits undesirable and redundant repetitions at various levels, such as token-level and paragraph-level \citep{dinan2019secondconversationalintelligencechallenge}.

Although the cause of repetition in LLMs is still not fully understood, some scholars have proposed methods to mitigate repetition. \citet{su2022contrastive} introduced the decoding method of contrastive search, which encourages diversity while maintaining the coherence of the generated text.
\citet{li2023repetition} demonstrate that penalizing repetitions in the training data significantly alleviates the degeneration problem in neural text generation. 
\citet{fu2021theoretical} presents a rebalanced encoding approach to address the issue of high inflow, reducing repetitions in both translation and language modeling tasks. However, the internal mechanisms of LLMs when they produce repetitive outputs remain insufficiently explored \citep{vaidya-etal-2023-humans}. \citet{zippert2020finding} proposed a psychological linkage between repetition and mathematics, and we want to find out if the similar repeat pattern mechanisms exist in the LLM. 

\noindent {\bf Language Model Mechanistic Interpretability.}
Mechanistic interpretability (MI) focuses on understanding the inner workings of neural networks, aiming to provide detailed insights into their computation processes and behavior \citep{bereska2024mechanistic,zhang2025eap,rai2024practicalreviewmechanisticinterpretability,zhang2024locate,zhang-etal-2025-mechanistic,hong2024dissecting,hu2024understanding}.
One approach to MI is the use of the logit lens \citep{nostalgebraist2020logitlens}, which focuses on interpreting what the model believes after each layer by examining the distributions generated by layers' activations. This approach allows us to observe how the model's predictions evolve and refine over the course of processing.

Another approach is to examine the features. Features are the things a network would ideally dedicate a neuron to if you gave it enough neurons \citep{Olahbases}. Researchers have developed sparse autoencoders (SAEs), which could decompose the activation into human-interpretable features \citep{sharkey2022superposition, cunningham2023sparse}. This process, known as sparse dictionary learning, reconstructs activation vectors as sparse linear combinations of directed vectors in the activation space \citep{bricken2023monosemanticity}.

Based on SAE, activation patching emerges as a method for further probing the role of individual features within a neural network. \citet{templeton2024scaling} demonstrated how steering the activation of the ``Golden Gate Bridge'' feature could influence the model to generate outputs specifically related to the Golden Gate Bridge, even when given diverse input prompts.


To develop LLMs, gaining mechanistic insights into their internal workings could reduce many risks \citep{nanda2022list}. Mechanistic interpretability enhances the predictability of future systems and reduces risks associated with deception and a foundation for model evaluation \citep{casper2023Interpretability}, bringing new perspectives to alignment work \citep{Ruthenis2023weneed}. \citet{geva2020transformer} argues that some components of the model (like the features) will catch repeating patterns, and the corresponding values tend to produce the same content in the output distribution, which may cause the model to produce repeated output. Through our work, we propose a solution to prevent the repetition problem, which can improve the performance of QA services and other text generation tasks.
\section{Sparse Autoencoders}
Sparse Autoencoders (SAEs) provide us with an approximate decomposition of the model’s activations into a linear combination of “feature directions” (SAE decoder weights) with coefficients equal to the feature activations. The sparsity penalty ensures that, for any given inputs to the model, only a small fraction of features will have nonzero activations. Thus, for any given token in any given context, the model activations are “explained” by a small set of active features (out of a large pool of possible features). Here’s how we perform this decomposition for activation $x$:
\begin{equation}
\hat{x} = b_{dec} + \sum_{i=1}^{F} f_i(x) W_{dec, i}.
\end{equation}
 The sum runs over all \( F \) features, effectively combining them to form the approximation of the original activation.  
Here \( \hat{x} \) is the reconstructed model activation, \( b_{dec} \in \mathbb{R}^D \) represents the learned bias term, \( W_{dec, i} \in \mathbb{R}^D \) are the learned decoder weights, and \( f_i(x) \) denotes the activation of the \( i \)-th feature, i.e., 
\begin{equation}
f(x) = \text{ReLU}(W_{enc} \cdot x + b_{enc})
\end{equation}
where \( f_i(x) \) is computed by passing the input \( x \) through the encoder weights \( W_{enc, i} \in \mathbb{R}^{F \times D} \) and the bias term \( b_{enc} \in \mathbb{R}^F \), followed by the ReLU nonlinearity. The ReLU function ensures that only positive activations are passed through, enforcing sparsity. The objective function encourages the model to maintain a sparse representation by minimizing the number of active features, which is defined as 
\begin{equation}
L(x) = \|x - \hat{x}\|_2^2 + \beta S(f_i(x)) +\alpha L_{aux}, 
\end{equation}
where $S$ is a function of the latent coefficients that penalize non-sparse decompositions (such as $\ell_1$ regularization), and $\beta$ is a sparsity coefficient. Some architectures also require the use of an auxiliary loss $L_{aux}$~\citep{gao2024scaling}.  

\paragraph{Steering with SAE.}
Steering is a method that utilizes the latent representations learned by SAE to steer the behavior of a model. In this process, the original activation is adjusted by introducing a steering coefficient, which controls the model's behavior. Specifically, the adjustment process can be expressed as: 
\begin{equation}\label{equ:steer}
\hat{X} = X + \lambda \cdot W_{\text{dec}}[\text{feature\_idx}]
\end{equation}
where $X$ represents the original activations tensor, $\hat{X}$ represents the modified activations tensor after steering, 
$\lambda$ is the steering coefficient, and $W_{\text{dec}}[\text{feature\_idx}]$ denotes the decoder weight vector corresponding to the steered feature index.

\section{Method}
In the following sections, we introduce the pipeline of casting ``Duplicatus Charm'' (DUC): (1) Repeat Pattern Construction; (2) Evaluation and Selection of Metrics; (3) Repetition Feature Identification; (4) Feature Steering.  
See  Figure \ref{fig:method} for the method overview.

\vspace{-0.1in}
\subsection{Repeat Pattern Construction\label{par:dataset}}
\vspace{-0.1in} 
As we mentioned, a challenge in identifying repeat features is developing an evaluation metric. To achieve this, we need to prepare a dataset with repetitions. However, to the best of our knowledge, currently, there is no readily available open-source dataset specifically designed for repetition tasks. To fill in the gap, we begin by constructing a custom repetition dataset. 
Based on previous work on analyzing repetition~\citep{altmann2015forms}, we particularly examine two forms of LLMs' repetition output: (a)  Token Repetition with excessive token-level recurrence where specific words/phrases replicate beyond natural language conventions and (b) Paragraph Repetition with structural redundancy through duplicated paragraph patterns.

We selected Orca-Chat\footnote{\url{https://huggingface.co/datasets/shahules786/orca-chat}}\citep{Orca-Chat}, a commonly used chat dataset containing short QA pairs, as our raw data. By applying specific rules to the output portions of this dataset, we can generate the desired repetition dataset. Specifically, 
we sampled 1,000 raw data, and the generated dataset consists of 5,500 samples. Among these, 4,500 belong to the token repetition scenario, and 1,000 belong to the paragraph repetition scenario. 

\noindent {\bf Token Repetition Scenario.}
In this scenario, we mainly generated repeated data based on two factors: 
$N$ is the token position where the repetition starts; $M$ is the number of tokens in the repeated token group. Dataset generation can be expressed as (\ref{equ:tokenrepetition}).

\begin{figure*}
\begin{equation}\label{equ:tokenrepetition}
\text{Token Repetition}(N,M) = \left( t_1, t_2, \dots, t_N, \underbrace{t_{N+1}, t_{N+2}, \dots, t_{N+M}}_{\text{repeated group}}, \underbrace{t_{N+1}, t_{N+2}, \dots, t_{N+M}}_{\text{repetition continues}} \right)
\end{equation}
\end{figure*}




We generated the dataset using 
$N$ values from an arithmetic sequence ranging from 0 to 140 with a common difference of 10 and $M$ values of 1, 2, and 5. Each case contains 100 dialogue samples, so in total, we have 4,500 dialogue samples.

\noindent {\bf  Paragraph Repetition Scenario.}
In this scenario, the entire text repeats continuously rather than just a few words. We generate the whole text five times to obtain the paragraph repetition texts. For this scenario, we sampled 1,000 raw data and applied repetition modifications.

\subsection{Repeat Curse Metric Selection\label{par:metric}}
Based on the dataset obtained in \S \ref{par:dataset}, we could evaluate the level of repetition using the difference metrics, ultimately selecting those that demonstrate discriminative capability for both scenarios. While \citep{li2023repetition} selected n-gram as the evaluation metric. However, we noticed that they did not clarify which value of $n$ performed best and whether there were better metrics. Here, we test different $n$ and introduce two additional potential metrics for comparison. The evaluation framework adopts two complementary approaches: The first set of metrics directly quantifies the degree of textual repetition, while the second approach conversely assesses the information content across the entire text. We have selected the following metrics for their effectiveness in addressing both dimensions: 


\paragraph{$n$-gram~\citep{li2023repetition}} The weighted repetition rate \( R \) is calculated as the ratio of the weighted sum of repeated \( n \)-grams to the maximum possible weighted sum:
\begin{equation}
    R = \frac{ \sum_{i \in n} f_i^{w} \quad \text{if} \ f_i > 1 }{ \sum_{i \in n} \max(f_i, 1)^{w} },
\end{equation} where \( n \) is the set of unique \( n \)-grams, \( f_i \) is the frequency of \( n_i \), and \( w \) is the weight factor. The numerator sums \( f_i^w \) for \( f_i > 1 \), while the denominator sums \( \max(f_i, 1)^w \) for all \( n \)-grams.

\paragraph{Self-BLEU \citep{papineni2002bleu}} BLEU score was originally used to evaluate machine translation performance. In this paper, we calculate the BLEU score of each sentence segment with other segments to obtain the average Self-BLEU score of the entire text, thereby evaluating the degree of repetition.
    $\text{Self-BLEU} = \frac{1}{n} \sum_{i=1}^{n} p_1(t_i),$ 
 where \( n \) is the total number of texts and \( p_1(t_i) \) is the 1-gram precision of text \( t_i \), calculated as the ratio of matching 1-grams to the total 1-grams in \( t_i \).

\paragraph{Information Entropy \citep{tsai2008information}} Since sentence lengths vary, we use maximum entropy for normalization:
\begin{equation}
    H_{\text{normalized}} = \frac{-\sum_{i=1}^{N} p_i \log_2(p_i)}{\log_2(N)}.
\end{equation}

\begin{figure*}[htbp]
    \centering
    \subfloat[n-gram: $M$ = 1]{\includegraphics[width=0.33\textwidth]{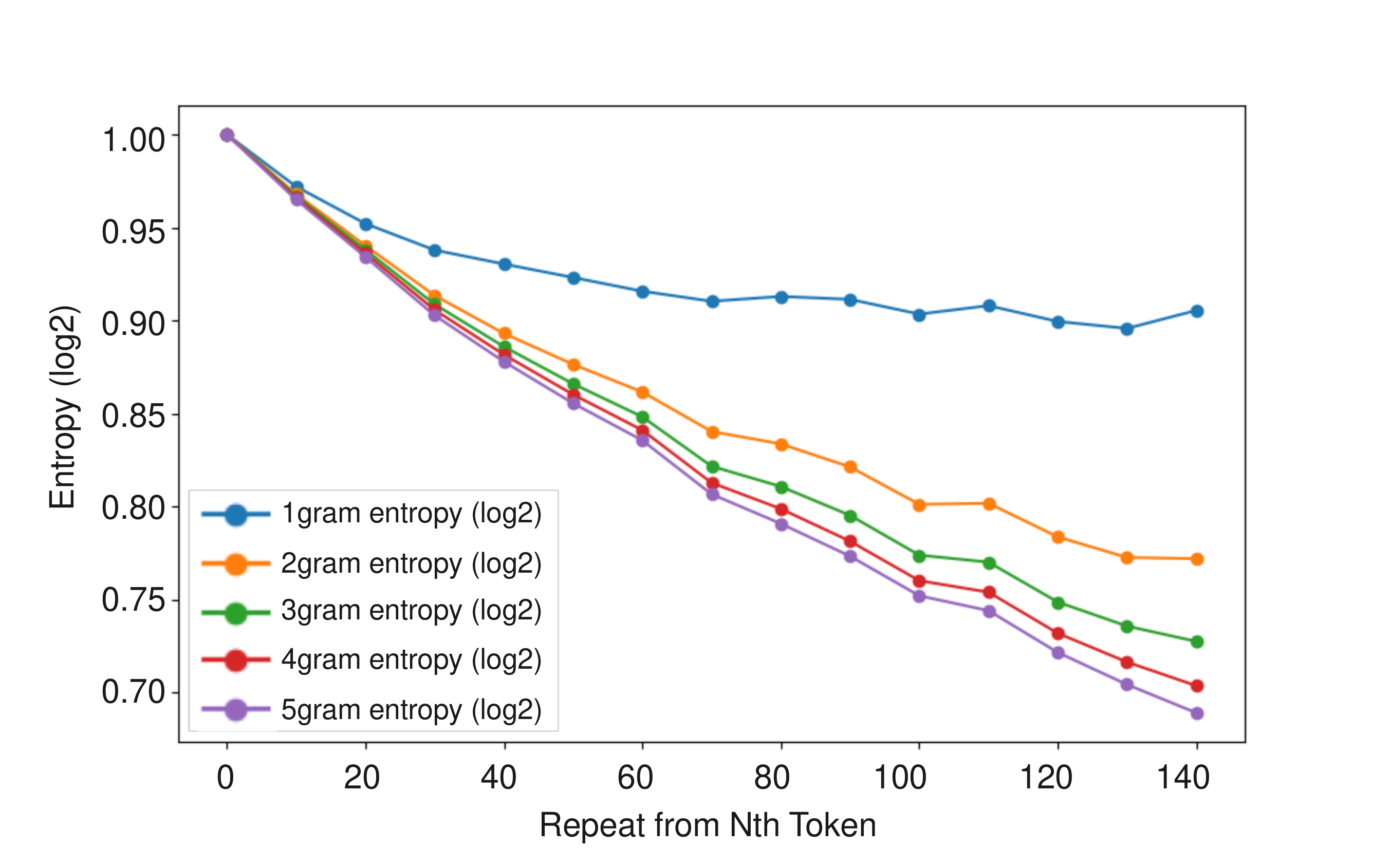}} 
    \subfloat[Entropy: $M$ = 1]{\includegraphics[width=0.33\textwidth]{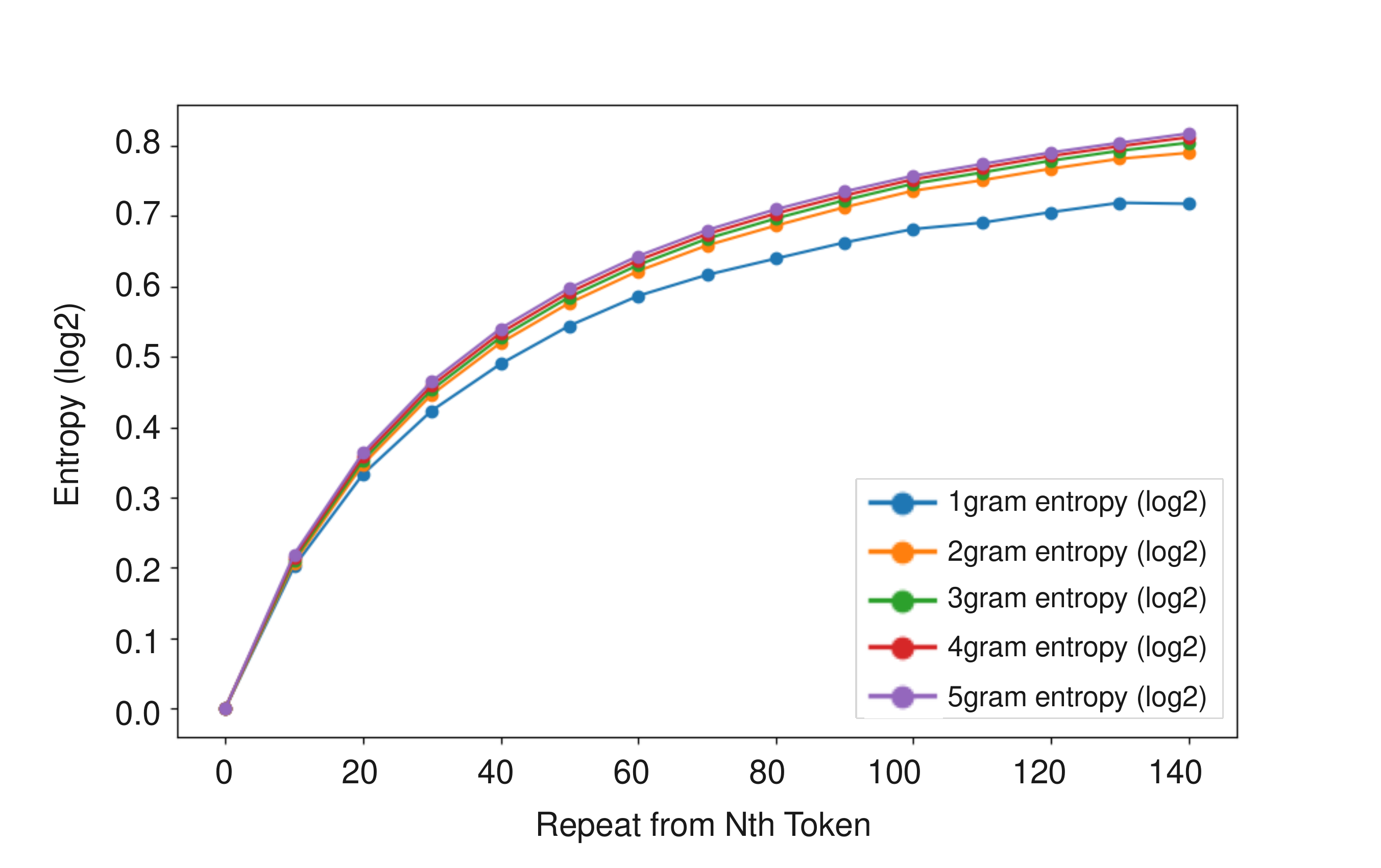}} 
    \subfloat[BLEU: $M$ = 1]{\includegraphics[width=0.33\textwidth]{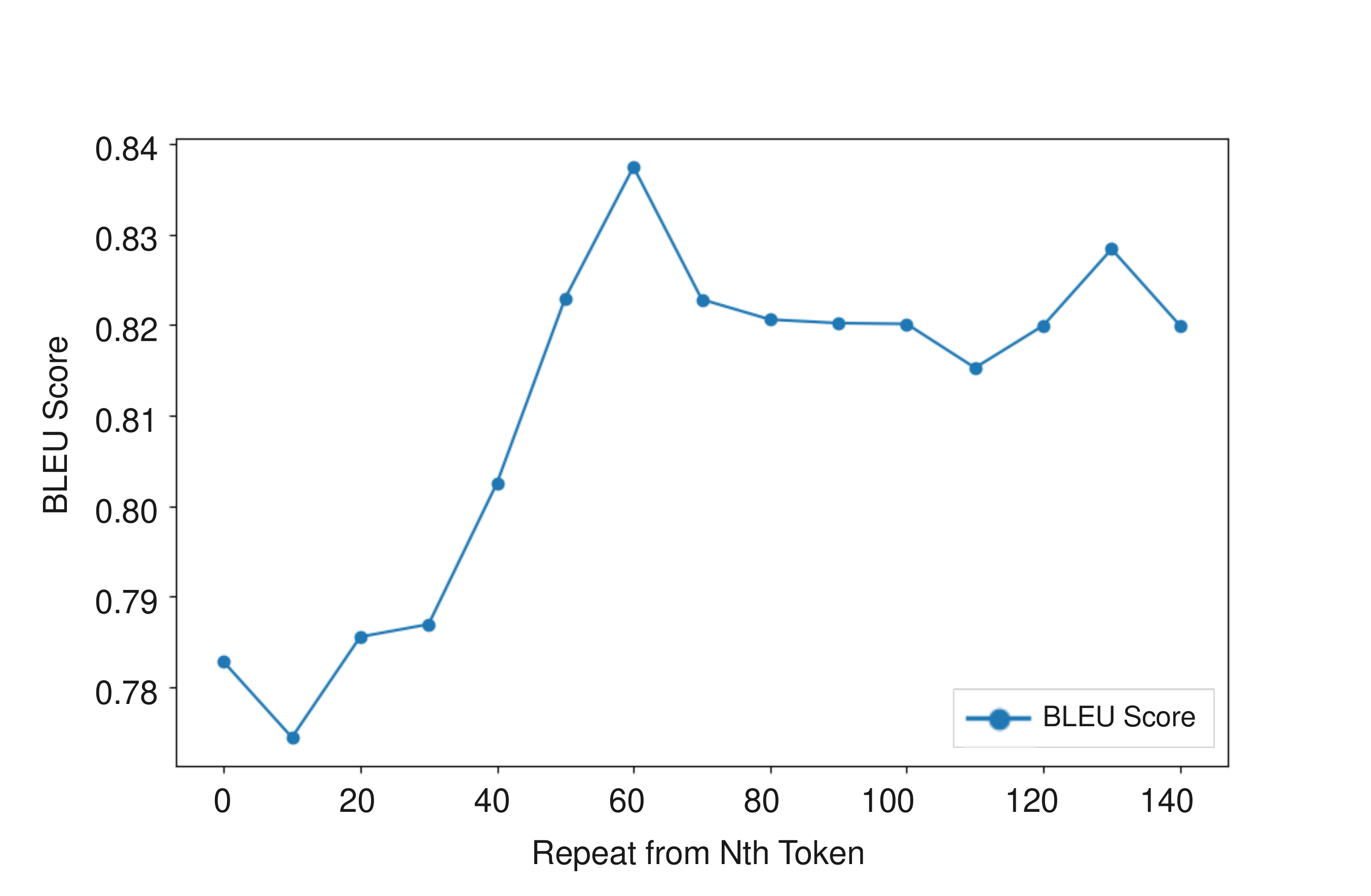}} 
    \caption{Comparison of Metrics in Token Repetition Scenario (M=1)}
    \label{fig:metrics}
\end{figure*}

   
    

\noindent {\bf Results of the Token Repetition Scenario}
In Figure \ref{fig:metrics}, we can see the information entropy curve for 1-gram differs from that of 2, 3, 4, and 5-grams, reaching its lowest value of around 0.6 when repeating from the 140-th token. This indicates that the optimal parameter choice for Information Entropy in this task is 1-gram, which provides strong distinguishability. Similarly, the 1-gram curve shows significant distinction compared to 2, 3, 4, and 5-grams, and can still accurately locate repetition situations above 0.9 when repeating from the 140-th token. 
The self-BLEU fluctuates within a difference of 0.1 when N takes different values, showing low distinguishability and poor performance.

Therefore, both the n-gram and information entropy metrics perform well with $n=1$. The rest results ($M=2,5$) are shown in Appendix \ref{app:eva metrics}.

\paragraph{Results of the  Paragraph Repetition Scenario}

In Figure \ref{fig:whole_paragraph_vs_original}, we can see the information entropy has a gap of 0.4 when evaluating the repetition and original data, while for n-gram the gap is 0.95, and it is 0.1 for BLEU. The comparison results show that the n-gram is highly sensitive in this scenario. However, when $n$ is set to 2, 3, or 4, we can also see the scores for normal text are too low, which is not conducive to subsequent analysis. Thus, $1$-gram has the best performance.




\subsection{Repeat Features Identification\label{par:identification}}
To identify effective repetition features, it is necessary to first locate the layers that contribute the most to the repetition phenomenon to narrow the search scope. Therefore, this section is divided into two steps: layer localization and feature localization.

\subsubsection{Layer Localization\label{sec:layerlocalization}}
Inspired by \citet{wang2022interpretability} 's pipeline, to determine the most important features, we decompose the residual stream and calculate the logit difference between the ''correct'' and ''incorrect'' answers.  In our problem, the next token that repeats the last token is considered ``correct'', while any token that does not repeat the last token is considered ``incorrect''.
In our work, given the input ``He hit Jack Jack Jack Jack Jack'', the correct output is ``Jack'', and it will be incorrect otherwise.  The layer with the largest logit difference is identified as the repetition layer.

Logit difference measures the difference in logit value between the two tokens, where a positive score means the correct token has a higher probability. 
In our work, given the input ``He hit Jack Jack Jack Jack Jack'', the correct output is ``Jack'', and the incorrect output is ``Jackson''. 
By calculating the difference between these two tokens, we can quantify the model's preference for the correct answer.
The formula for the logit difference direction is given by:
\begin{equation}
    \ell_{\text{diff\_direction}} = 
    c_{\text{direction}}- i_{\text{direction}},
\end{equation}
where $c_{\text{direction}}$ and $i_{\text{direction}}$ represent the residual stream directions for the correct and incorrect answers, respectively. 




Finally, we calculate the layer attribution by taking the dot product of the residual activations at each layer with the previously computed logit difference direction, $\ell_{\text{diff\_direction}}$. This operation quantifies the contribution of each layer to the final prediction. The formula for the logit contribution at layer $\ell$ is: 
\begin{equation}
\ell_{\text{contribution}_\ell}= \text{residual}_{\ell} \cdot \ell_{\text{diff\_direction}}, 
\end{equation}
where $\text{residual}_{\ell}$ is the residual activation.


Based on the work of \citet{wang2022interpretability}, who utilized 10 templates to locate indirect object layers, we adopted a similar approach tailored to our work. We create 8 templates with induced repeated generation inputs (refer to Table \ref{tab:prompts_answers}). Appendix \ref{app:layerattribution} shows that the contributions of the intermediate layers and the final layer to generating repeated content are the most significant. Therefore, we will look for repetition features in both the intermediate and final layers.

\subsubsection{Feature Localization\label{par:featurelocalization}}
Through \textbf{\S \ref{sec:layerlocalization}}, we will further localize the feature on the most significant layer and the second most significant layer~\citep{wang2022interpretability}.


We employ a pre-trained SAE model of each model, which has already captured meaningful features. Then by setting the features' steering coefficient $\lambda$ in (\ref{equ:steer}) as 1.5-2 times the original activation level, we were able to enhance the content related to the generated features without causing model collapse, which refers to the failure of the model to generate meaningful or diverse outputs, caused by disrupting the balance of the model's parameters and structure~\citep{arena3.0}.

Based on the generated text after activation, we determine that features with repeat score (RS) (See \textbf{\S \ref{par:metric}}) above $\rho$ are considered repetition features.

\begin{equation*}
\text{Feature} = 
\begin{cases} 
\text{Repetition Feature} & \text{if } \text{RS} \geq \rho \\
\text{Common Feature} & \text{if } \text{RS} < \rho.
\end{cases}
\end{equation*}


\section{Experiment}
\subsection{Setup}

\noindent {\bf Models}
We specifically selected large pre-trained models that have open-sourced their SAE models: GPT2-small \citep{gpt2} with GPT-sm-res-jb \citep{gptsae}; Gemma-2-2B \citep{gemma_2024} with GemmaScope-res-16k \citep{gemmasae}; Llama-3.1-8B \citep{llama3} with LlamaScope-res-32k \citep{llamasae}.

\noindent {\bf Datasets and Metric}
We used three datasets: two containing hard (academic) and easy questions, and one containing intuitively easy to repeat enumeration questions. We selected the Academic ShortQA~\footnote{\url{https://huggingface.co/datasets/DisgustingOzil/Academic_dataset_ShortQA}}\citep{DisgustingOzil2024} (AQ), which contains hard (academic) questions, and Natural Questions~\footnote{\url{https://huggingface.co/datasets/google-research-datasets/natural_questions}} \citep{kwiatkowski-etal-2019-natural} (NQ), which contains simple questions.We selected diversity enumeration problems from the Diversity-Challenge-Dataset\footnote{\url{https://huggingface.co/datasets/YokyYao/Diversity_Challenge} \citep{YokyYao2025}} (EQ). Compared with ordinary problems, these are more challenging. We randomly select some of the questions shown in the Appendix \ref{app:EQ}.


Following the result of \textbf{\S\ref{par:metric}}, we use n-gram as the repeat score to evaluate the degree of the repeat curve.

\noindent {\bf Hyperparameters} In our method, we have two hyperparamaters $\rho$ and $\lambda$.  We will set $\rho=0.4$, which is based on human evaluations (refer to Appendix \ref{app:humanevaluation}). And we set  $\lambda=2$, which this value ensures that the model's overall performance remains unaffected while strongly inducing the occurrence of the Repeat Curse~\citep{arena3.0}.

\begin{table*}[!htbp]
\centering
\resizebox{\textwidth}{!}{
\begin{tabular}{lccccccccccccccc}
\toprule
\rowcolor{blue!10}
& \textbf{Dataset} & \multicolumn{4}{c}{\textbf{Activation Ratio}} & \textbf{Dataset} & \multicolumn{4}{c}{\textbf{Activation Ratio}} & \textbf{Dataset} & \multicolumn{4}{c}{\textbf{Activation Ratio}} \\
\cmidrule(lr){1-16} 
\rowcolor{blue!10}
\textbf{Model and Layer} & EQ & 10\% & 20\% & 50\% & 100\% & AQ & 10\% & 20\% & 50\% & 100\% & NQ & 10\% & 20\% & 50\% & 100\% \\
\midrule

\multirow{4}{*}{GPT2-small Layer 9} & original & 0.37 & 0.37 & 0.37 & 0.37 & original & 0.25 & 0.25 & 0.25 & 0.25 & original & 0.18 & 0.18 & 0.18 & 0.18\\
 & activated(CF) & 0.36 & 0.37 & 0.37 & 0.37 & activated(CF) & 0.25 & 0.25 & 0.26 & 0.27 & activated(CF) & 0.18 & 0.19 & 0.19 & 0.21 \\
 & activated(RF) & 0.55 & 0.60 & 0.68 & \textbf{0.72} & activated(RF) & 0.50 & 0.53 & 0.55 & 0.58 &
 activated(RF) & 0.47 & 0.52 & 0.54 & 0.55  \\
 & deactivated & 0.33 & 0.32 & 0.21 & 0.19 & deactivated & 0.25 & 0.25 & 0.23 & 0.23 & deactivated & 0.18 & 0.18 & 0.16 & 0.17\\
\midrule
\multirow{4}{*}{GPT2-small Layer 11} & original & 0.35 & 0.35 & 0.35 & 0.35 & original & 0.25 & 0.25 & 0.25 & 0.25 & original & 0.19 & 0.19 & 0.19 & 0.19 \\
 & activated(CF) & 0.35 & 0.35 & 0.36 & 0.37 & activated(CF) & 0.25 & 0.26 & 0.25 & 0.27 & activated(CF) & 0.19 & 0.19 & 0.19 & 0.21 \\
 & activated(RF) & 0.53 & 0.58 & 0.66 & \textbf{0.70} & activated(RF) & 0.50 & 0.50 & 0.51 & 0.53 & activated(RF) & 0.45 & 0.46 & 0.49 & 0.51 \\
 & deactivated & 0.34 & 0.30 & 0.23 & 0.22 & deactivated & 0.25 & 0.25 & 0.24 & 0.24 & deactivated & 0.18 & 0.19 & 0.19 & 0.18 \\
\midrule
\multirow{4}{*}{Gemma-2-2B Layer 22} & original & 0.28 & 0.28 & 0.28 & 0.28 & original & 0.22 & 0.22 & 0.22 & 0.22 & original & 0.19 & 0.19 & 0.19 & 0.19 \\
 & activated(CF) & 0.28 & 0.28 & 0.30 & 0.32 & activated(CF) & 0.22 & 0.22 & 0.24 & 0.24 & activated(Cf) & 0.19 & 0.19 & 0.19 & 0.21 \\
 & activated(RF) & 0.51 & 0.56 & 0.64 & \textbf{0.68} & activated(RF) & 0.41 & 0.44 & 0.45 & 0.48 & activated(RF) & 0.38 & 0.43 & 0.44 & 0.48   \\
 & deactivated & 0.25 & 0.25 & 0.24 & 0.25 & deactivated & 0.22 & 0.22 & 0.21 & 0.20 & deactivated & 0.19 & 0.19 & 0.19 & 0.19 \\
\midrule
\multirow{4}{*}{Gemma-2-2B Layer 24}& original & 0.29 & 0.29 & 0.29 & 0.29 & original & 0.24 & 0.24 & 0.24 & 0.24 & original & 0.20 & 0.20 & 0.20 & 0.20 \\
 & activated(CF) & 0.29 & 0.30 & 0.31 & 0.33 & activated(CF) & 0.25 & 0.25 & 0.25 & 0.27 & activated(CF) & 0.20 & 0.21 & 0.21 & 0.22 \\
 & activated(RF) & 0.49 & 0.54 & 0.61 & \textbf{0.65} & activated(RF) & 0.42 & 0.44 & 0.47 & 0.52 & activated & 0.42 & 0.44 & 0.46 & 0.49 \\
 & deactivated & 0.36 & 0.27 & 0.24 & 0.20 & deactivated & 0.24 & 0.24 & 0.26 & 0.24 & deactivated & 0.20 & 0.17 & 0.18 & 0.17 \\
\midrule
\multirow{4}{*}{Llama-3.1-8B Layer 24} & original & 0.28 & 0.28 & 0.28 & 0.28 & original & 0.21 & 0.21 & 0.21 & 0.21 & original & 0.19 & 0.19 & 0.19 & 0.19 \\
 & activated(CF) & 0.28 & 0.29 & 0.29 & 0.29 & activated(CF) & 0.21 & 0.21 & 0.21 & 0.23 & activated(CF) & 0.20 & 0.19 & 0.19 & 0.20 \\
 & activated(RF) & 0.46 & 0.50 & 0.57 & \textbf{0.62} & activated(RF) & 0.40 & 0.41 & 0.44 & 0.46 & activated(RF) & 0.36 & 0.37 & 0.41 & 0.43 \\
 & deactivated & 0.24 & 0.25 & 0.19 & 0.19 & deactivated & 0.21 & 0.19 & 0.19 & 0.19 & deactivated & 0.19 & 0.18 & 0.18 & 0.17\\
\midrule
\multirow{4}{*}{Llama-3.1-8B Layer 29} & original & 0.25 & 0.25 & 0.25 & 0.25 & original & 0.21 & 0.21 & 0.21 & 0.21 & original & 0.15 & 0.15 & 0.15 & 0.15 \\
 & activated(CF) & 0.25 & 0.26 & 0.27 & 0.27 & activated(CF) & 0.21 & 0.20 & 0.24 & 0.26 & activated(CF) & 0.19 & 0.19 & 0.20 & 0.20 \\
 & activated(RF) & 0.48 & 0.52 & 0.60 & \textbf{0.66} & activated(RF) & 0.39 & 0.40 & 0.44 & 0.45 & activated(RF) & 0.36 & 0.36 & 0.37  & 0.39\\
 & deactivated & 0.25 & 0.26 & 0.19 & 0.18 & deactivated & 0.20 & 0.20 & 0.19 & 0.21 & deactivated & 0.15 & 0.16 & 0.14 & 0.14 \\
\bottomrule
\end{tabular}
}
\caption{Effect of Repetition Feature Activation at Different Levels (10\%, 20\%, 50\%, 100\%). We take experiments on 3 datasets: Enumeration Questions (EQ), Academic Questions (AQ), Natural Questions (NQ).  ``CF'' refers to randomly selected common feature, and ``RF'' refers to repetition feature. \textbf{Bold} indicates the highest score of each model\label{tab:conclusion}}
\end{table*}

\subsection{Main Result}
\noindent {\bf{Repetition Features}}
This part will present the repetition features identified based on different datasets and analyze their characteristics.
We iterated through each feature of the repetition layer, activated them, randomly sampled questions from each dataset to query the model, and used the repeat score to evaluate the generated results to identify repetition features. All the identified repetition features are shown in Appendix \ref{app: repeatfeature}. 

\begin{figure}[htbp]
\centering
\includegraphics[width=0.5\textwidth]{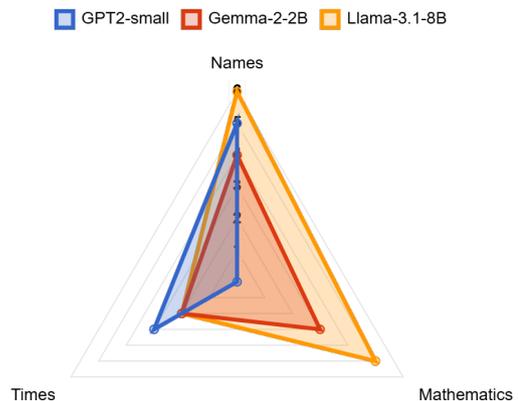}  
\caption{Illustration of the distribution of characteristics for repetition features identified two or more times across multiple datasets.}  
\label{fig:radar} 
\end{figure}

We find that the repetition features identified two or more times across the three datasets are associated with Names, Time, and Mathematics (see Figure \ref{fig:radar}). The model that identified the same repetition feature the most is Llama-3.1-8B, while the least is GPT2-small. We did not identify the same mathematics-related feature in GPT2-small, which reflects its instability in mathematical reasoning. Overall, among the repetition features identified from the three models, names are the most likely to cause repetition.

\noindent {\bf{Evaluation of DUC}}
We activate the repetition feature in batches at each layer of each model, analyze the repeat score of the generated results, and evaluate whether the DUC is effective. Next, we attempted to reduce the steering coefficient of these features to see whether it can mitigate Repeat Curse. We perform experiments on 3 datasets (EQ, AQ, NQ), sequentially activating 10\%, 20\%, 50\%, and 100\% of the repetition features (the most influential ones are prioritized). After multiple trials, we calculated the average repeat score for the generated text. The detailed results are presented in Table \ref{tab:conclusion}.

The activated common features (CF) serve as the baseline for the study. After activating an equal number of common features as repetition features, there was no significant change in repeat scores.

From the dataset perspective of view, the repeat score is highest on the EQ dataset, followed by a gradual decrease on the AQ and NQ datasets. This indicates that questions that induce repetition exhibit a more severe Repeat Curse when activating repetition features. Regarding the difficulty of the questions, the more challenging the question, the more pronounced the Repeat Curse becomes after activation. 

After deactivating the repetition feature, the repeat score for the EQ dataset shows the most significant change, while the scores for AQ and NQ exhibit only minor fluctuations around their original values, occasionally even exceeding them (e.g., Gemma-2-2B Layer 24 Activation Ratio=50\%). This indicates that the EQ dataset, which originally had a higher score, is more sensitive to the deactivation of the repetition feature, resulting in a larger difference. This suggests that the effectiveness of mitigating Repeat Curse relies on the presence of a certain degree of inherent repetition in the problem itself. Without this foundational repetition, the impact of such measures may not be observable.


For layers,  the ones that contribute more significantly tend to achieve higher repeat scores. For instance, in GPT2-small Layer 9, which has a greater contribution, consistently yield higher scores across all three datasets under ``activated repetition feature (RF)'' compared to the 11th layer.

For models, GPT2-small exhibited the highest repeat score after activation, with a range of approximately 0.6. This indicates that GPT2-small has a higher sensitivity to repetition features, whereas larger models like Gemma-2-2B and Llama-3.1-8B are more robust to mitigate such effects.

\noindent {\bf{Mitigation Effect} }
To further investigate the practical value of DUC, we also need to evaluate its effectiveness in mitigating the Repeat Curse. We introduced Information Entropy as a positive metric complementary to the Repeat Score and employed Perplexity to assess whether the generative capability of the deactivated model was affected.


\begin{table*}[htbp]            
\centering          
\scalebox{0.60}{
\begin{tabular}{@{}lllllllllllll@{}}           
\toprule          
\multirow{2}{*}{\textbf{Dataset}} & \multirow{2}{*}{\textbf{Method}} &\multicolumn{3}{c}{\textbf{Gemma-2-2B}} &\multicolumn{3}{c}{\textbf{Llama-3.1-8B}} & \multicolumn{3}{c}{\textbf{GPT2}} \\    
\cmidrule(lr){3-5} \cmidrule(lr){6-8} \cmidrule(lr){9-11}

  & & $\mathrm{Repeat Score}\downarrow$ & $\mathrm{Entropy}\uparrow$ & $\mathrm{Perplexity}\downarrow$ & $\mathrm{Repeat Score}\downarrow$ & $\mathrm{Entropy}\uparrow$ & $\mathrm{Perplexity}\downarrow$ & $\mathrm{Repeat Score}\downarrow$ & $\mathrm{Entropy}\uparrow$ & $\mathrm{Perplexity}\downarrow$ & \\        
\midrule            
\multirow{5}{*}{EQ} & 
Greedy & 0.28 & 0.86 & \textbf{8.53} & 0.26 & \textbf{0.88} & 16.72 & 0.36 & 0.81 & 8.70\\      
& Beam & 0.31 & 0.84 & 11.13 & 0.22 & 0.89 & \textbf{ 9.23} & 0.46 & 0.70 & \textbf{7.66} \\   
& TopK & 0.14 & 0.94 & 24.37 & 0.19 & 0.91 & 10.46 & 0.13 & \textbf{0.95} & 30.47\\   
& TopP & 0.14 & 0.94 & 16.74 & 0.19 & 0.91 & 13.32 & 0.19 & 0.94 & 74.05 \\   
 
 & \textbf{DUC (ours)} & \textbf{0.13\down{0.18}} & \textbf{0.94\up{0.1}} & 17.23 & \textbf{0.13\down{0.13}} & 0.91 & 10.39 & \textbf{0.13\down{0.33}} & 0.94 & 23.14\\  
 
\cmidrule(l){1-11}  

\multirow{5}{*}{AQ} & Greedy & 0.22 & 0.89 & 54.14 & 0.21 & 0.90 & 14.62 & 0.25 & 0.88 & 8.70\\      
& Beam & 0.35 & 0.84 & 85.40 & 0.52 & 0.76 & \textbf{ 8.60} & 0.53 & 0.75 & \textbf{6.88}\\   
& TopK & 0.16 & 0.93 & 61.79 & 0.36 & 0.84 & 11.89 & 0.21 & 0.91 & 31.00 \\   
& TopP & 0.17 & 0.93 & 60.79 & 0.35 & 0.84 & 10.42 & 0.17 & 0.93 & 55.27 \\   
 
 & \textbf{DUC (ours)} & \textbf{0.10\down{0.25}} & \textbf{0.96\up{0.12}} & 
 \textbf{ 8.84\down{76.56}} & \textbf{0.11\down{0.41}} & \textbf{0.93\down{0.17}} & 
 11.17 & 
 \textbf{0.14\down{0.39}} & \textbf{0.94\up{0.19}} &
 14.53 & 
 \\      
 
\cmidrule(l){1-11}     

\multirow{5}{*}{NQ} & Greedy & 0.18 & 0.92 & \textbf{5.02} & 0.19 & 0.91 & 4.98 & 0.18 & 0.91 & 7.41\\      
& Beam & 0.47 & 0.77 & 87.64 & 0.36 & 0.82 & \textbf{3.56} & 0.51 & 0.71 & \textbf{4.39} \\   
& TopK & 0.20 & 0.91 & 65.52 & 0.27 & 0.87 & 4.93 & 0.16 & 0.93 & 33.16\\   
& TopP & 0.20 & 0.91 & 66.53 & 0.25 & 0.88 & 5.35 & \textbf{0.13} & 0.93 & 82.81 \\   
  
 & \textbf{DUC (ours)} & \textbf{0.11\down{0.36}} & \textbf{0.95\up{0.18}} & 
 14.72 & 
 \textbf{0.11\down{0.25}} & \textbf{0.95\up{0.13}} & 
  3.97 & 
  0.14 & 
  \textbf{0.94\up{0.23}} &
 22.51 \\      
\bottomrule       
\end{tabular}}
   
\caption{\textbf{Mitigation results}. We conducted experiments on three datasets, and all models were tested across five methods. \textbf{Bold} indicates the highest score. \label{tab:chair}} 
\end{table*}

The results in Table \ref{tab:chair} show that DUC demonstrates nearly the best performance in reducing repetition across various models and datasets. Both the Repeat Score and Information Entropy achieved an 88.9\% state-of-the-art (SOTA) probability, with no significant decrease in Perplexity. Notably, the Perplexity of Gemma-2-2B on the AQ dataset even increased instead of decreasing, further validating the practical utility of the DUC method. 

And we conduct two case studies to further investigate the \textbf{Utility After Mitigation} in Appendix \ref{app:New}.

\noindent {\bf{Visualization Results} }
In Appendix \ref{app:D}, Table \ref{tab:20199} and Table \ref{tab:100} respectively show the effects of feature activation on repetition features and regular features before and after activation. In Table \ref{tab:20199}, feature 20199 directly causes a Repeat Curse. In Table \ref{tab:100}, feature 100 represents words related to political campaigns and candidates, and its generation after steering consistently includes references to ``president''.

Table \ref{tab:controlrepeatcurse} provides an example of the output results under the condition where 100\% of the repetition features are deactivated, offering a clear demonstration of the mitigation.

\section{Conclusion}
In this paper, we take a perspective from the feature level and introduce a pipeline named ``Duplicatus Charm'' (DUC). Through this mechanistic interpretability method, we can identify the repetition features. By activating the repetition feature, we can induce the Repeat Curse, which was then evaluated through repeat scores and validated by humans in our experiment. And by deactivating these features will mitigate the Repeat Curse. Furthermore, we summarize the common characteristics of repetition features across three models. 
\section{Limitations}
It is worth mentioning that there are still several limitations in this study.


\paragraph{Repeat Score}
The identification of repetitive features relies on a predefined threshold for the repeat score ($\rho=0.4$), which was determined based on human evaluation. This introduces a potential for subjectivity, as different threshold choices could lead to different sets of repetitive features.

\paragraph{Models}
The experiments were conducted on three LLMs with pre-trained SAE (GPT2-small, Gemma-2-2B, and Llama-3.1-8B), which have relatively limited scales. 
\citet{he2024llamascope}, \citet{gao2024scaling}, \citet{cunningham2023sparse} mentioned the limitations of open-source SAEs and said most researches on SAE are using 7B or larger models. Moreover, due to the limited availability of open-source SAEs, we have done our best to refine this study. With the continuous development of the SAE community in the future, we will keep updating this work.

\section{Acknowledgement}
Di Wang and Shu Yang are supported in part by the  funding BAS/1/1689-01-01, URF/1/4663-01-01,  REI/1/5232-01-01,  REI/1/5332-01-01,  and URF/1/5508-01-01  from KAUST, and funding from KAUST - Center of Excellence for Generative AI, under award number 5940.
\bibliography{acl}
\clearpage
\appendix
\section{Evaluation of Metrics\label{app:eva metrics}}
\begin{figure*}
\centering
    \subfloat[Entropy: $M$ = 2]{\includegraphics[width=0.3\textwidth]{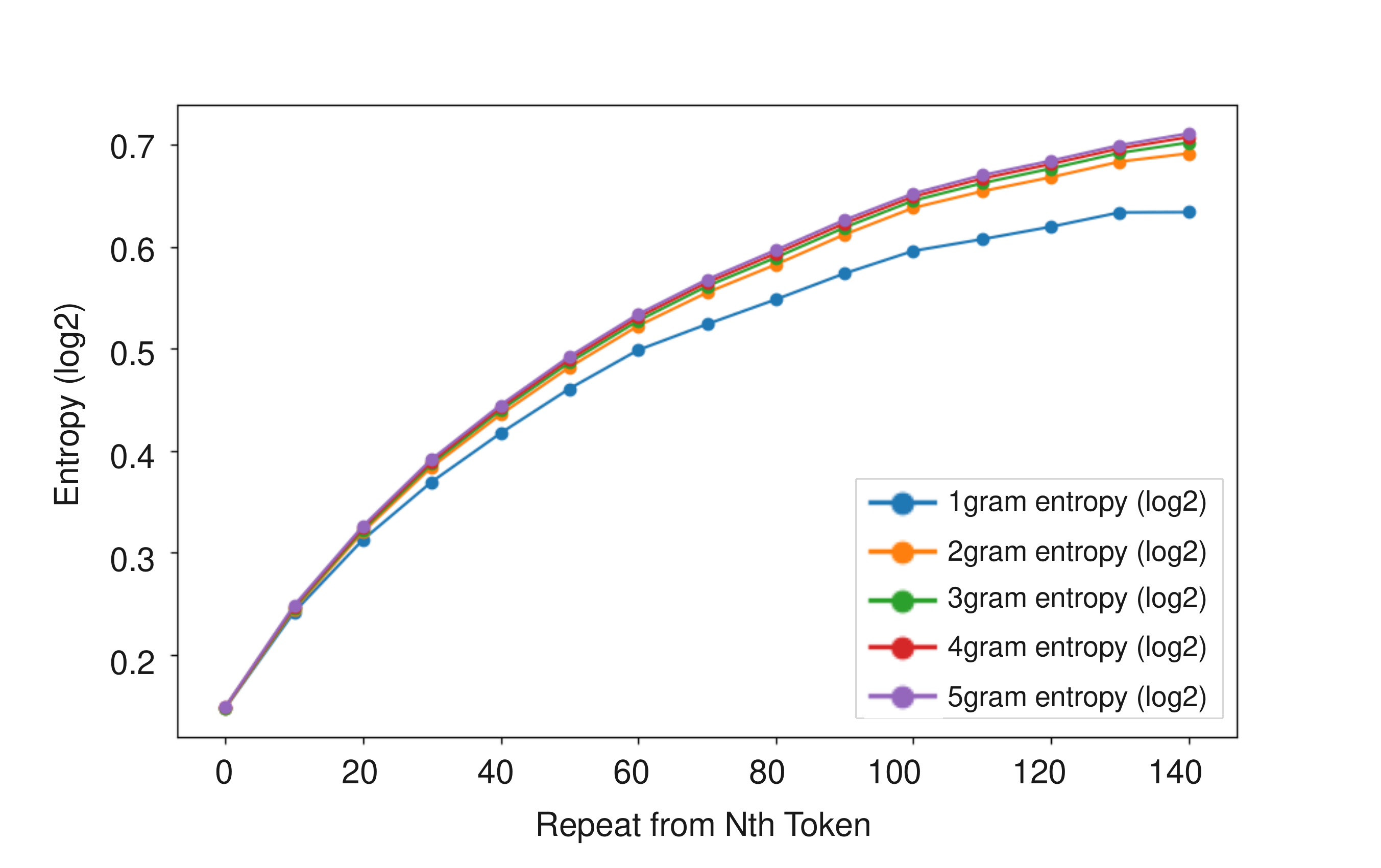}} 
    \subfloat[n-gram: $M$ = 2]{\includegraphics[width=0.3\textwidth]{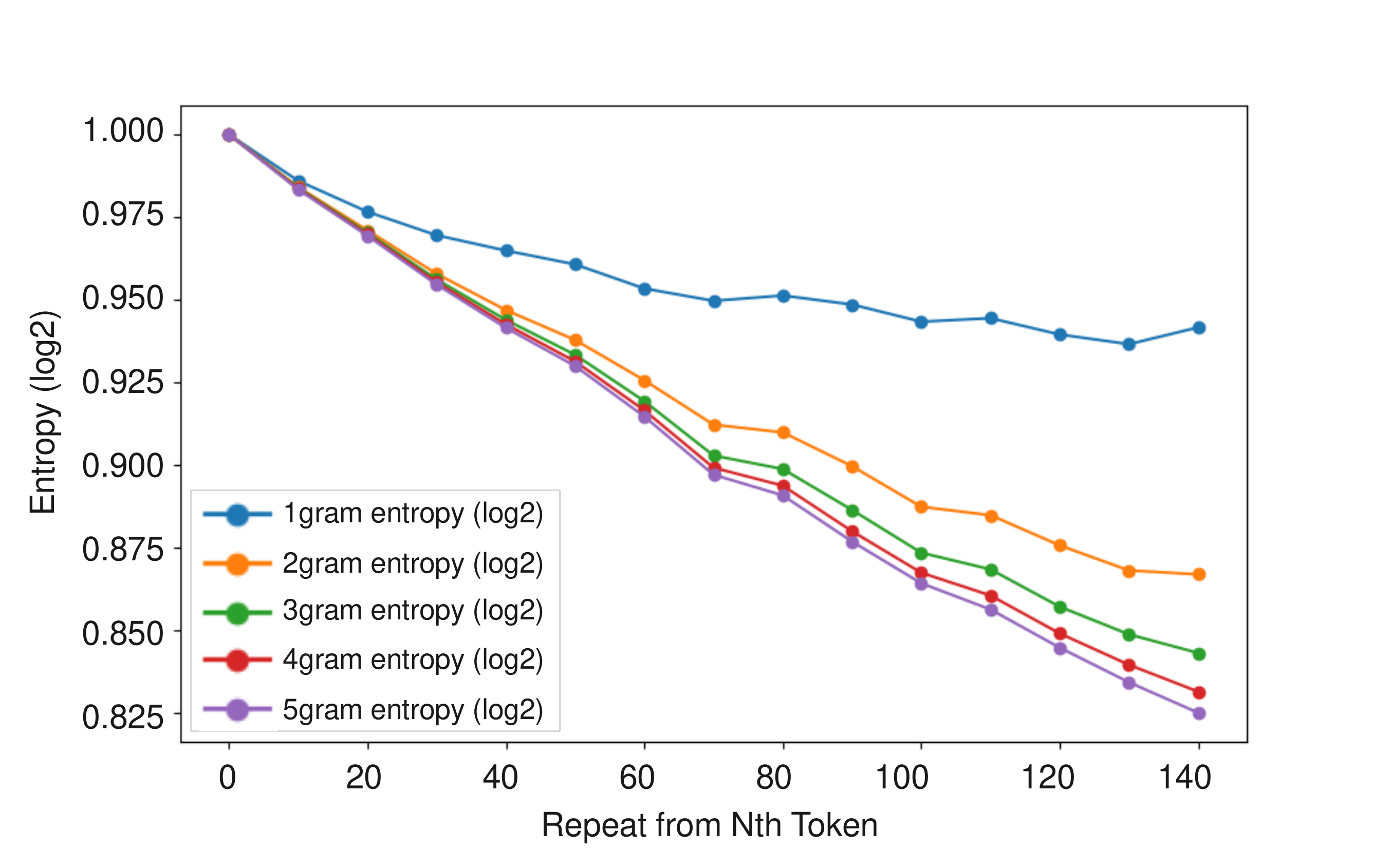}} 
    \subfloat[BLEU:$M$ = 2]{\includegraphics[width=0.3\textwidth]{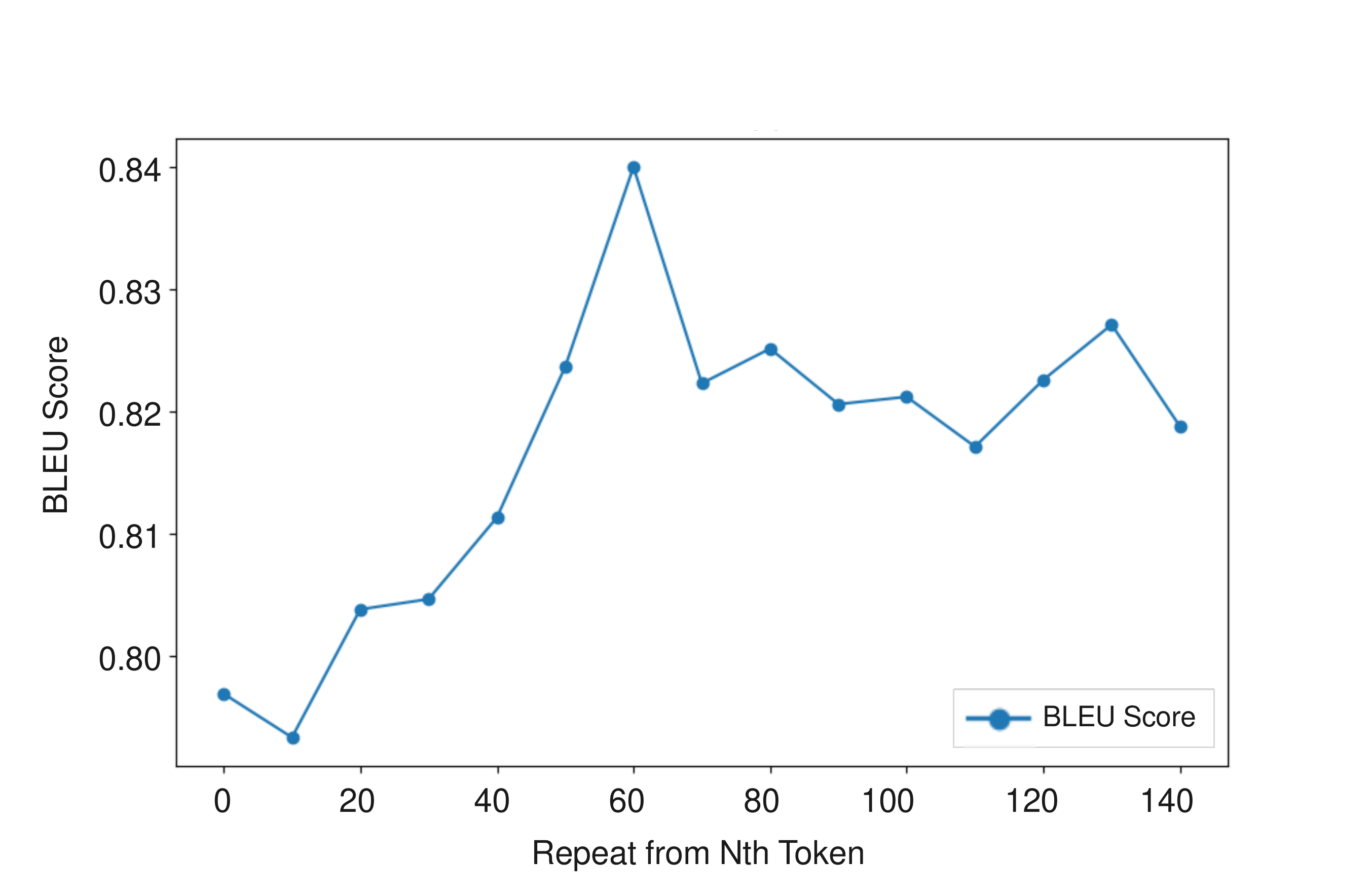}} \\
    \subfloat[Entropy: $M$ = 5]{\includegraphics[width=0.3\textwidth]{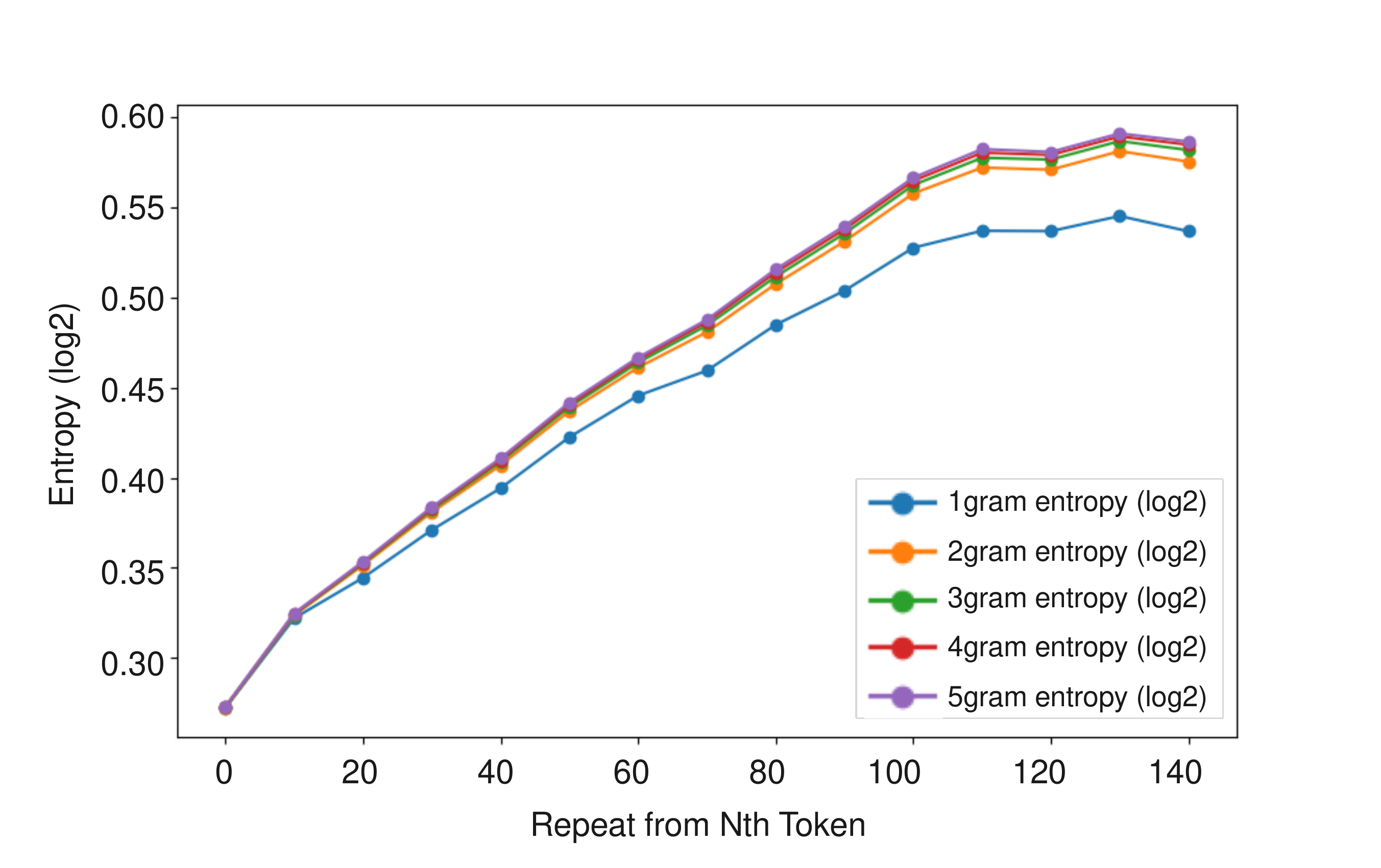}} 
    \subfloat[n-gram: $M$ = 5]{\includegraphics[width=0.3\textwidth]{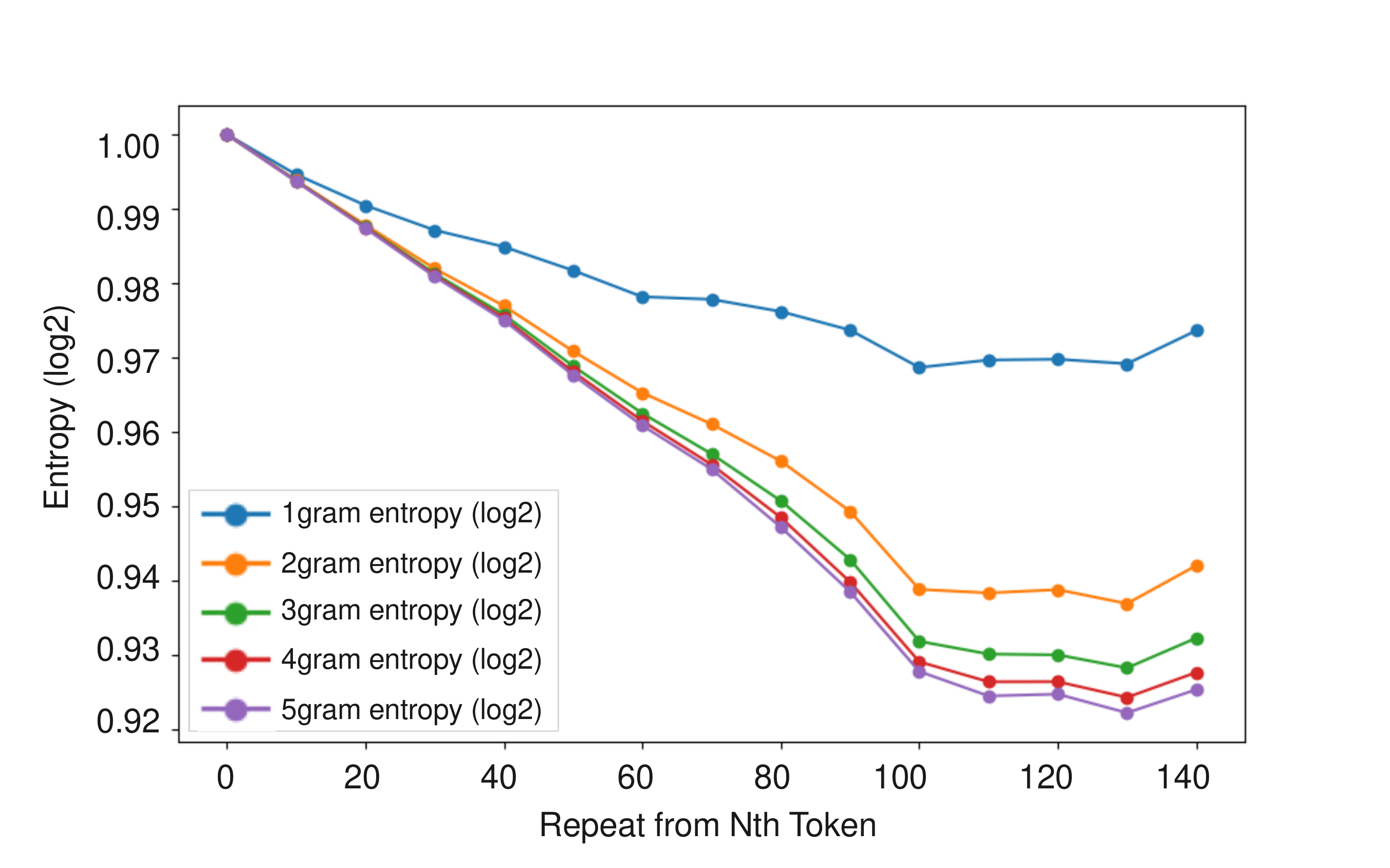}} 
    \subfloat[BLEU: $M$ = 5]{\includegraphics[width=0.3\textwidth]{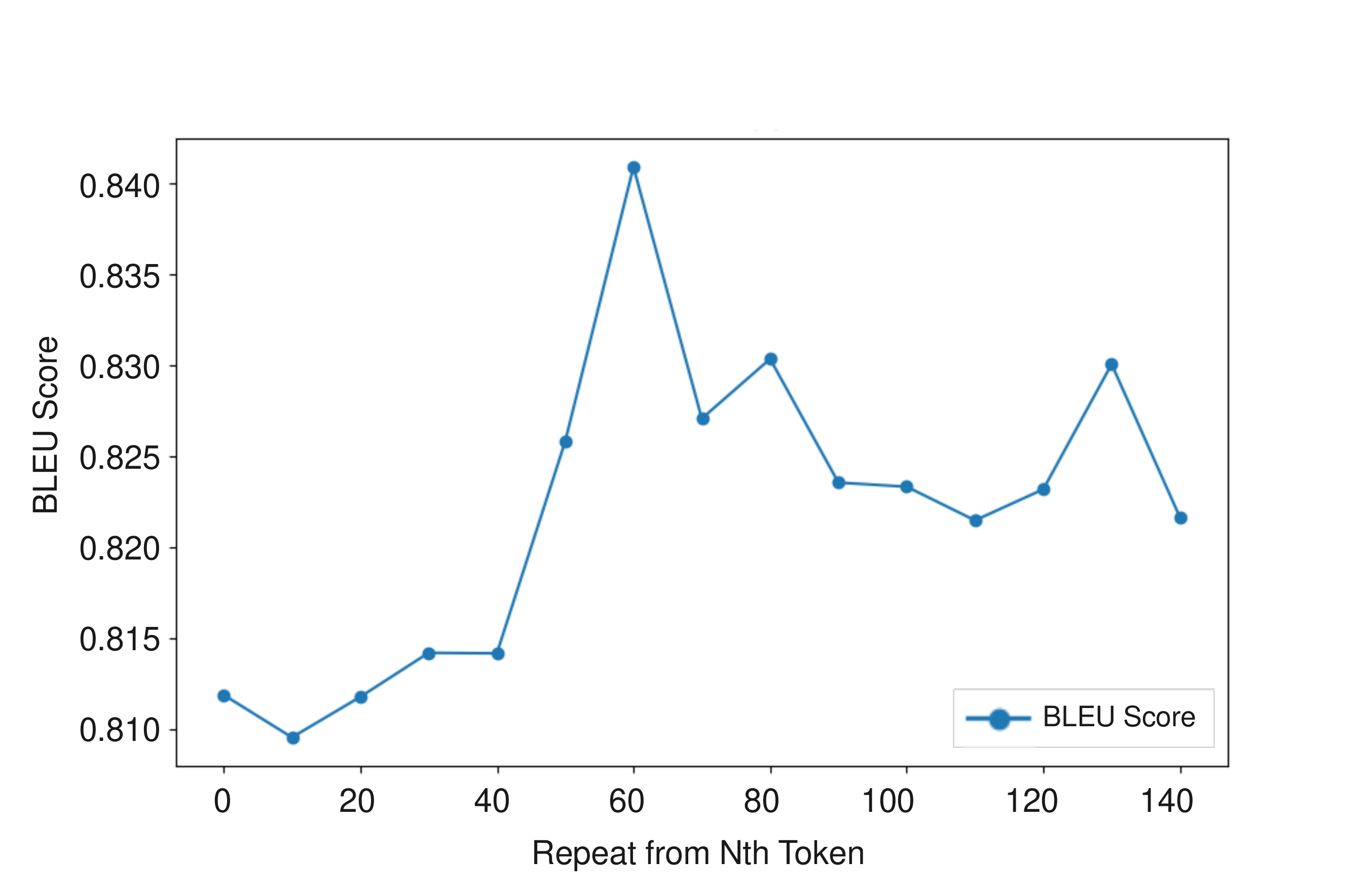}} 
    \caption{Comparison of Metrics in Token Repetition Scenario (M=2, 5) \label{fig:rest matrics}}
\end{figure*}

Information Entropy represents the amount of information contained, so when repeated positions occur later, more information is included, resulting in an upward trend in the curve. On the other hand, the n-gram directly describes the repeated content, so when the repeated positions occur later, the proportion of repeated content within the overall content becomes smaller, leading to a downward trend in the curve. Figure \ref{fig:rest matrics} shows the comparison result when $M=2,5$.

\begin{figure}[htbp]
\centering
\includegraphics[width=0.49\textwidth]{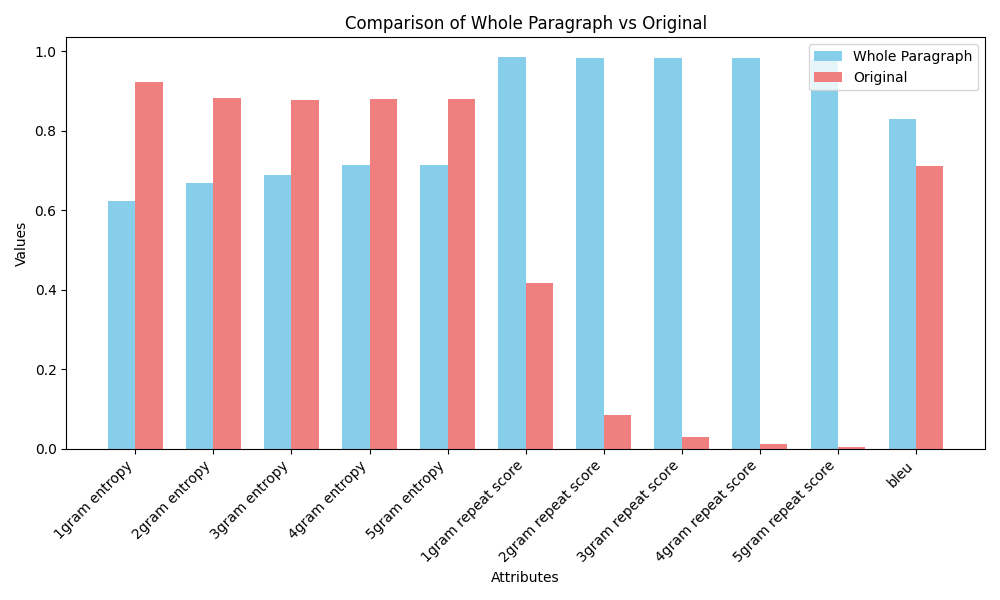}  
\caption{Comparison of Metrics in Paragraph Repetition Scenario}  
\label{fig:whole_paragraph_vs_original} 
\end{figure}
\section{Layer Attribution\label{app:layerattribution}}
\begin{figure}[htbp]
\centering
\includegraphics[width=0.5\textwidth]{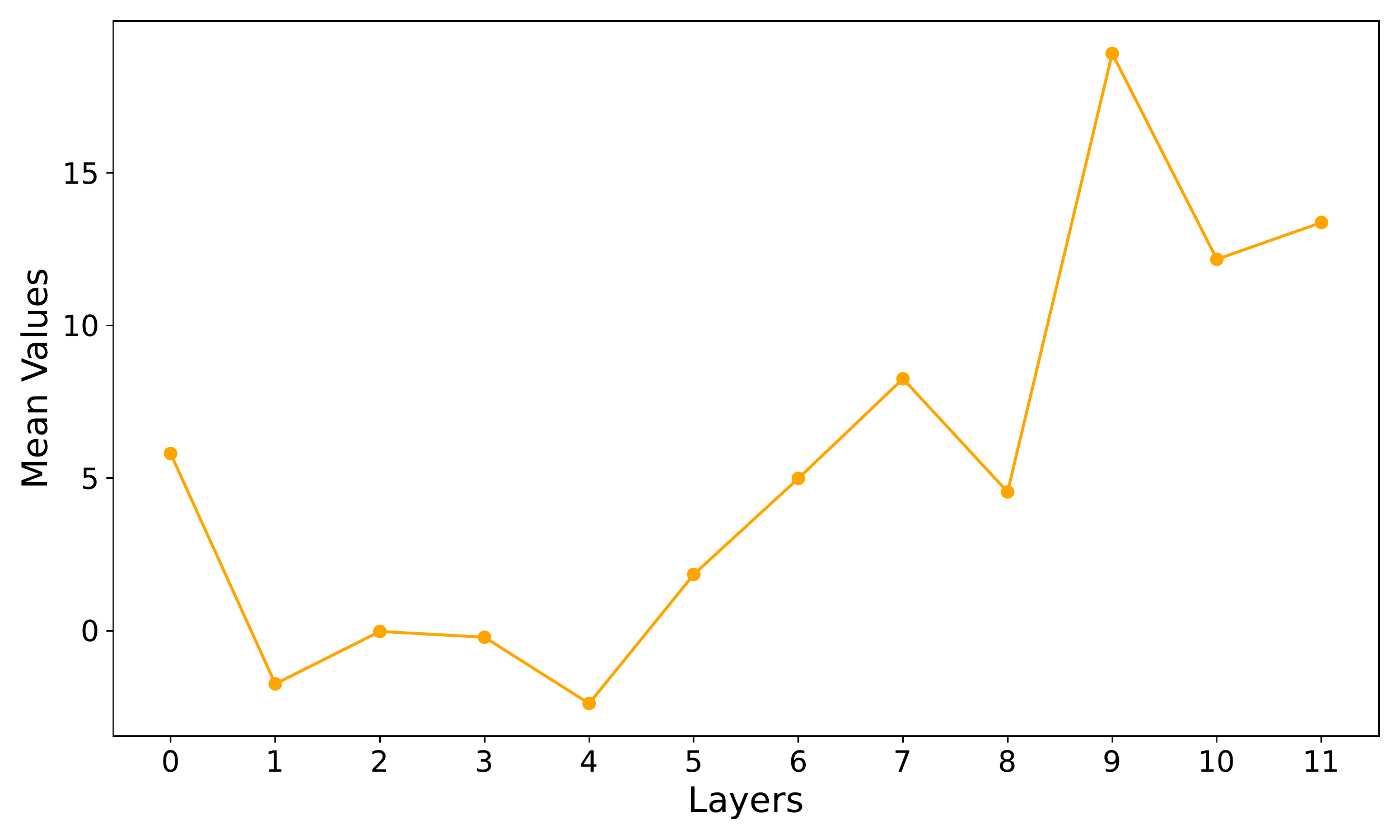}  
\caption{GPT2-small Layer Attribution}  
\label{fig:gpt_layer_attribution} 
\end{figure}

\begin{figure}[htbp]
\centering
\includegraphics[width=0.5\textwidth]{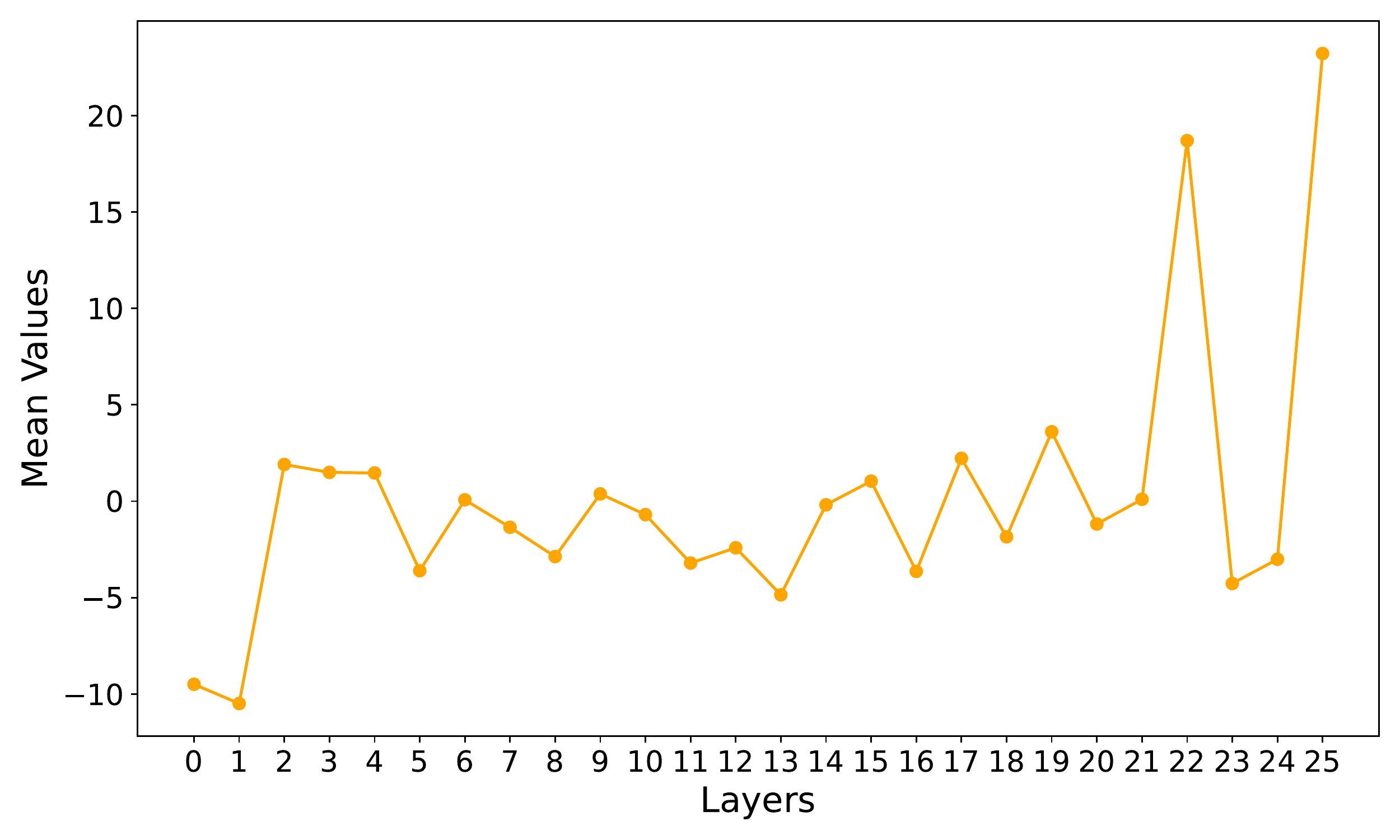}  
\caption{Gemma-2-2B Layer Attribution}  
\label{fig:gemma_layer_attribution} 
\end{figure}

\begin{figure}[htbp]
\centering
\includegraphics[width=0.5\textwidth]{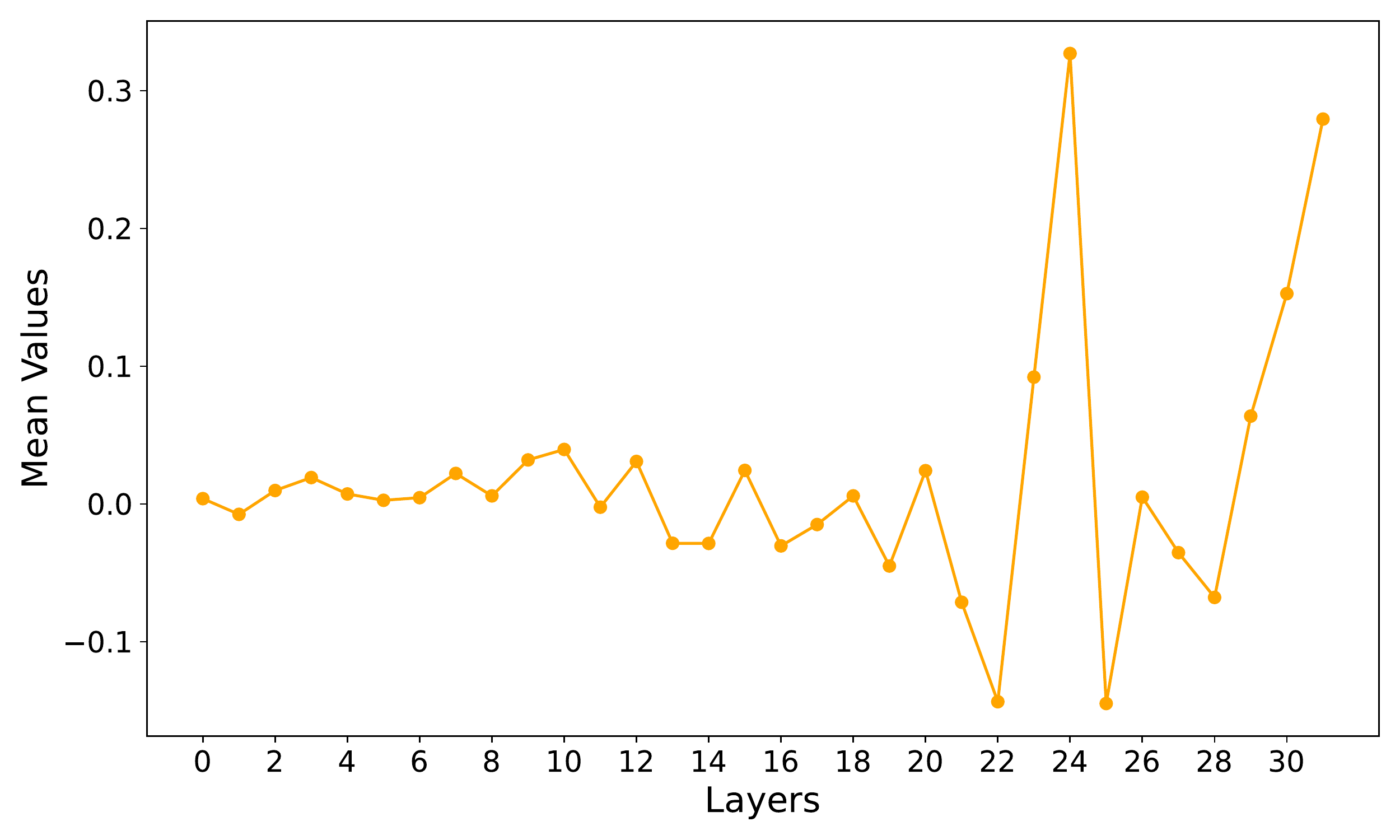}  
\caption{Llama-3.1-8B Layer Attribution}  
\label{fig:llama_layer_attribution} 
\end{figure}

We provide the eight templates of prompts and answers used to investigate the influence of layer contributions on repetition. Each prompt was designed to include repeated tokens at specific intervals to induce patterns of repetition. The answers were defined by selecting tokens at the corresponding positions in the prompt as “correct” when they were the same as the previous token and “incorrect” when they differed. 
We recorded the residual difference direction used to measure the difference between 'correct' and 'incorrect' generation and further quantified the contribution of each layer to the final prediction by calculating the dot product of each layer's activations and the residual difference direction. The results are shown in Figure \ref{fig:gpt_layer_attribution}, \ref{fig:gemma_layer_attribution} and \ref{fig:llama_layer_attribution}.
The prompts and answers are in Table \ref{tab:prompts_answers}.
\begin{table*}[htbp]
\centering
\resizebox{\textwidth}{!}{
\begin{tabular}{|c|c|}
\hline
\textbf{Prompt} & \textbf{Answer(Correct, Incorrect)} \\
\hline
\texttt{Is displacement is a vector or scalar is a vector is is a vector is is} & \texttt{a, vector} \\
\texttt{School school school is a place where you school school school is a school is a school is a} & \texttt{school, place} \\
\texttt{Which does not has an index does not has an index} & \texttt{does, and} \\
\texttt{Friends friends friends are people who friends friends friends are people who friends are people who} & \texttt{friends, help} \\
\texttt{Speed speed speed is a scalar that speed speed speed is a speed is a speed is a} & \texttt{speed, scalar} \\
\texttt{Mass mass mass does not change mass mass mass changes doesn't change} & \texttt{mass, anything} \\
\texttt{Work done is energy is energy is} & \texttt{energy, to} \\
\texttt{Time is always measured in seconds Time is always measured in seconds Time is always} & \texttt{measured, limited} \\
\hline
\end{tabular}
}
\caption{Tamplates of Induced Repeated Generation Inputs and Answers}
\label{tab:prompts_answers}
\end{table*}


\section{Repeat Feature\label{app: repeatfeature}}

We present the identified repetition features of the three models on three datasets in Table \ref{tab:gptfeatureeq}, \ref{tab:gptfeatureaq}, \ref{tab:gptfeaturnq}, \ref{tab:gemmafeatureeq}, \ref{tab:gemmafeatureaq}, \ref{tab:gemmafeaturenq}, \ref{tab:llamafeatureeq}, \ref{tab:llamafeatureaq}, \ref{tab:llamafeaturenq}. These include Layer 9 and Layer 11 of GPT2-small, Layer 22 and Layer 24 of Gemma-2-2B, and Layer 24 and Layer 29 of Llama-3.1-8B. The \underline{underline} indicates that the feature appears twice across the three models, while the \textbf{bold} indicates that the feature appears three times. For more detailed feature information, you can search the corresponding model's feature ID at \url{https://www.neuronpedia.org}. 

\begin{figure}[htbp] 
    \centering  \includegraphics[width=0.48\textwidth]{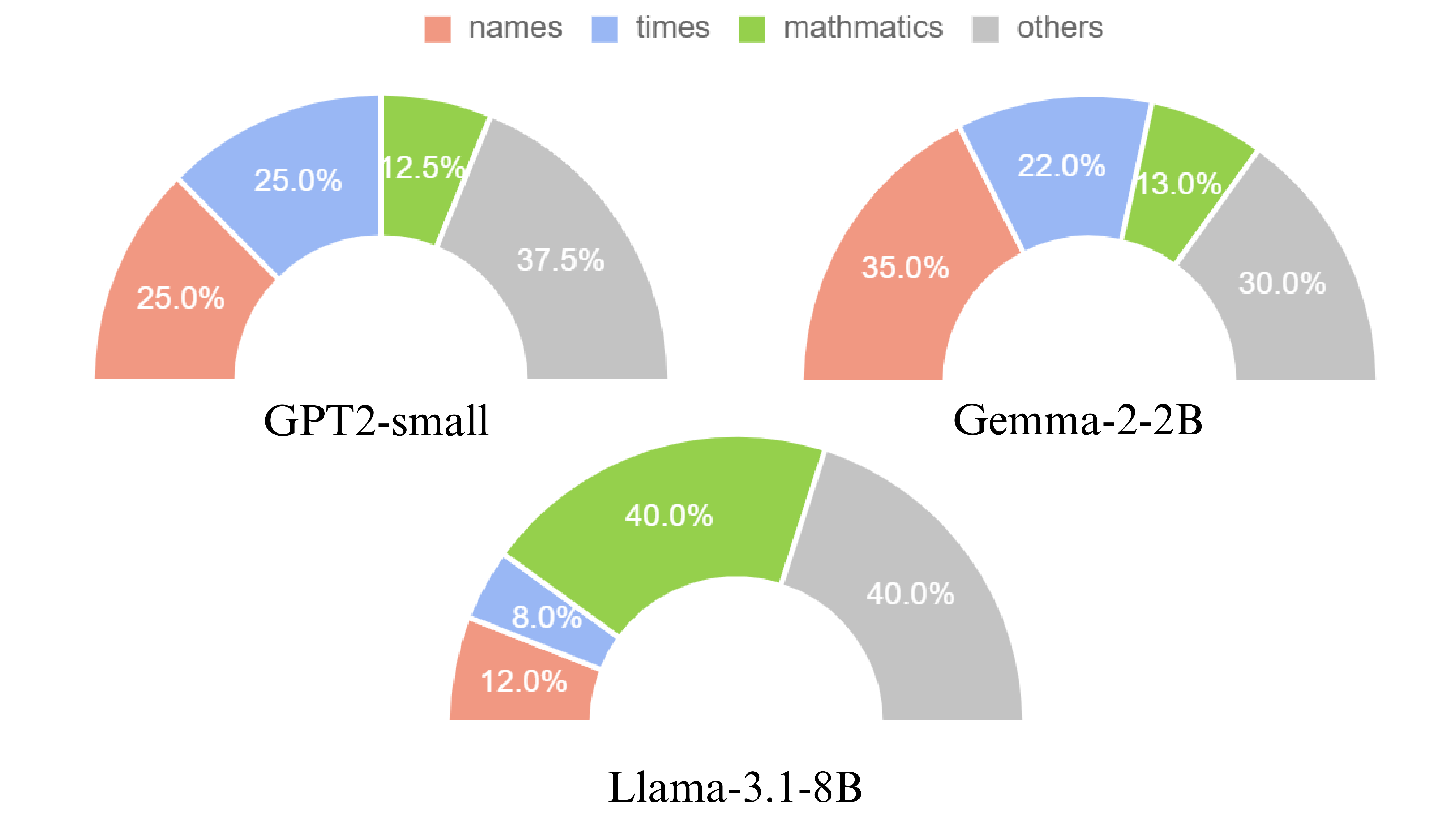}
    \caption{Feature Characteristic of Each Model on academic question (AQ) dataset.}
    \label{featurecharacteristic}
\end{figure}

In Figure \ref{featurecharacteristic}, we illustrate the distribution of feature characteristics in the AQ dataset, where Llama-3.1-8B demonstrates a significantly higher number of mathematics-related features compared to other models. 

\begin{table*}[ht]
\centering
\begin{tabular}{|c|p{12cm}|}
\hline
\rowcolor{red!10}
\textbf{Feature ID} &  \textbf{Description} \\
\hline 
\multicolumn{2}{|c|}{\textbf{Layer 9}} \\
\hline
6643 & periods and punctuation marksindicating the end of sentences\\
\underline{6972}  & \underline{entities such as names, organizations, and transferred amounts} \\
8700 & phrases related to popular moviefranchises and their connections \\
13299 & expressions related to deep emotionsand personal connections \\
\textbf{13944} & \textbf{proper names of individuals or entities} \\
\textbf{16888} & \textbf{names specifically with initials followed by periods} \\
17533 & information about pricing andsubscriptions \\
19200 & environmental elements such asocations including caves, mountainslakes, and specific physical objects \\
20161 & financial and economic data points orindicators \\
22587 & discount-related terms and actions \\
23516 &  mentions of names, specifically those related to the character Jack and others in a specific narrative context\\

\hline
\multicolumn{2}{|c|}{\textbf{Layer 11}} \\
\hline
\textbf{6023} & \textbf{proper names of individual} \\
7413 & phrases related to problem-solving and improvement\\
8860 & proper nouns and specific terms related to legal and politicalmatters\\
8919& phrases related to online security and encryption\\
10226 & words related to political figures or events \\
\underline{10431} & \underline{temporal references or expressions related to time} \\
\underline{11642} & \underline{locations and spatial references} \\ 
\textbf{12078} & \textbf{dates or events when something occurred} \\
13140 & references to specific numerical codes or identifiers\\
15084 & phrases related to personal evaluation or judgment \\
15405 & phrases related to negative events or experiences \\

\hline
\end{tabular}
\caption{GPT2-small Repetition Features (EQ)\label{tab:gptfeatureeq}}
\end{table*}

\begin{table*}[ht]
\centering
\begin{tabular}{|c|p{12cm}|}
\hline
\rowcolor{red!10}
\textbf{Feature ID} &  \textbf{Description} \\
\hline 
\multicolumn{2}{|c|}{\textbf{Layer 9}} \\
\hline
3615  & references to product offers and services, likely related to advertising or marketing content \\
3661  & numerical information related to accounting or distribution \\
\underline{6972}  & \underline{entities such as names, organizations, and transferred amounts} \\
7798  & names related to Middle Eastern politics and conflicts \\
8357  & information related to news articles and events, focusing on dates and locations \\
10178 & references to the television show "Game of Thrones" \\
\textbf{13944} & \textbf{proper names of individuals or entities} \\
16631 & policy-related phrases like "full employment," "de facto amnesty," "mass deportation," and "no-fly zone." \\
\textbf{16888} & \textbf{names specifically with initials followed by periods} \\
18380 & government department names and related entities \\
22275 & cities and locations \\
22317 & phrases related to keeping business operational or in progress \\

\hline
\multicolumn{2}{|c|}{\textbf{Layer 11}} \\
\hline
2868 & elements related to coding orprogramming concepts \\
3185 & locations expressed as intersectionsor addresses \\
\textbf{6023} & \textbf{proper names of individual} \\
6038  & words related to the name "Kris" \\
8353  & terms related to geographic locations or businesses \\
\underline{10431} & \underline{temporal references or expressions related to time} \\
\underline{11642} & \underline{locations and spatial references} \\ 
\textbf{12078} & \textbf{dates or events when something occurred} \\
18623 & terms related to financial capital and taxes \\
20971 & phrases indicating events or activities related to time and context \\
\underline{22640} & \underline{specific time-related events or processes} \\
23164 & measurement units and quantitiesrelated to mathematics and physics \\

\hline
\end{tabular}
\caption{GPT2-small Repetition Features (AQ)\label{tab:gptfeatureaq}}
\end{table*}

\begin{table*}[ht]
\centering
\begin{tabular}{|c|p{12cm}|}
\hline
\rowcolor{red!10}
\textbf{Feature ID} &  \textbf{Description} \\
\hline 
\multicolumn{2}{|c|}{\textbf{Layer 9}} \\
\hline
181 & information about who directed and wrote a film or TV show \\
1238 & phrases related to confidence and mental states \\
1554 & names or references to names in a text \\

3660 & references to an exchange of goods or services \\
4688 & phrases related to age groups \\
6792 & Roman numerals followed by letters and numbers \\
8047 & people or places associated with specific names\\
12969 & terms related to indexing, such as words like ``index'' and actions related to creating or comparing indexes \\
\textbf{13944} & \textbf{proper names of individuals or entities} \\

\textbf{16888} & \textbf{names specifically with initials followed by periods} \\
17290 & references to video games \\
19121 & political party names, such as AAP, Greens, Congress, and NDP, along with related terms\\
20636 & references to computer science concepts related to object-oriented programming \\

\hline
\multicolumn{2}{|c|}{\textbf{Layer 11}} \\
\hline
692 & references to individuals named or related to ``Bhutan''\\
3017 & names of people or entities preceded by a title or username \\
3299 & numbers and codes with a specific
structure\\
4464 & topics related to government, politics, and various industries \\
\textbf{6023} & \textbf{proper names of individual} \\
\textbf{12078} & \textbf{dates or events when something occurred} \\
16594 & Proper nouns,specifically names of people and locations \\
16765 & specific parts of objects or machines \\
17956 & technical terms related to geologyand physics \\
18371 & mentions of people's names in a social context \\
\underline{22640} & \underline{specific time-related events or processes} \\

\hline
\end{tabular}
\caption{GPT2-small Repetition Features (NQ)\label{tab:gptfeaturnq}}
\end{table*}

\begin{table*}[ht]
\centering
\begin{tabular}{|c|p{12cm}|}
\hline
\rowcolor{red!10}
\textbf{Feature ID} & \textbf{Description} \\
\hline
\multicolumn{2}{|c|}{\textbf{Layer 22}} \\
\hline
259 & references to the color red, particularly in varying contexts or phrases \\
2603 & mention of characters or entities named "Daika" along with their various attributes and relationships\\

\textbf{3509} & \textbf{names and titles of individuals in professional contexts} \\
7362 & mentions of Washington, D.C., and variations of its name \\
\underline{5327} & \underline{names or mentions of a specific individual or group} \\
7535 & terms and phrases associated with research and funding in the scientific field \\
8726 & terms and phrases associated with ``cross-linking'' concepts \\
11734 & symbols and variables related to math, physics, and statistics, particularly in the context of equations and mathematical notation \\
12235 & LaTeX math syntax related to mathematical symbols and expressions \\
\textbf{14137} & \textbf{phrases related to durations and periods of time} \\
\underline{15056} & \underline{certain key terms and phrases related to various subjects such as programming,} \\ 
& \underline{medicine, and science} \\[2pt]
\hline
\multicolumn{2}{|c|}{\textbf{Layer 24}} \\
\hline
\textbf{1119} & \textbf{numerical values or formats in mathematical or programming contexts} \\
2497 & references to sports leagues and tournaments \\
\underline{5333} & \underline{date ranges and time periods} \\
7923 & specific statistics related to baseball performance \\
\underline{8789} & \underline{numerical values, particularly those indicating ages or durations} \\
\textbf{11892} & \textbf{names of characters in a narrative context} \\
12519 & acronyms and specific terms related to molecular biology or chemistry \\
14510 & elements related to programming and data structures \\
14995 & mentions of the name ``Tom'' \\
16307 & information related to financial transactions and corporate activities \\

\hline
\end{tabular}
\caption{Gemma-2-2B Repetition Features (EQ)}\label{tab:gemmafeatureeq}
\end{table*}

\begin{table*}[ht]
\centering
\begin{tabular}{|c|p{12cm}|}
\hline
\rowcolor{red!10}
\textbf{Feature ID} & \textbf{Description} \\
\hline
\multicolumn{2}{|c|}{\textbf{Layer 22}} \\
\hline
\textbf{3509} & \textbf{names and titles of individuals in professional contexts} \\
3576 & dates and times related to events or records \\
5618 & phrases indicating the degree of proximity or likelihood, particularly words like "almost." \\
5947 & phrases that indicate long-term perspectives or considerations \\
9363 & references to dates or time-related
events \\
10278 & terms related to programming syntax and variable naming conventions \\
13028 & names of researchers and contributors involved in a project or study \\
14041 & keywords and phrases related to actions and intentions, particularly involving deception or retrieval \\
\textbf{14137} & \textbf{phrases related to durations and periods of time} \\
14370 & reterences to ages and years of experience \\

\hline
\multicolumn{2}{|c|}{\textbf{Layer 24}} \\
\hline
\textbf{1119} & \textbf{numerical values or formats in mathematical or programming contexts} \\
2505 & specific names and references to legal proceedings or court cases \\
4795 & phrases indicating social connections and personal interactions \\
7079 & names of contributors or authors associated with a research project \\
7158 & names and identities of notable individuals associated with Ballymena \\
7719 & terms related to subscription models and billing options \\
8137 & terms and phrases related to biochemical processes and treatments involving heavy metals or chemical interactions \\

8491 & technical terms and phrases related to programming or mathematicalconcepts \\

8777 & phrases related to expressions of gratitude and acknowledgments \\

\underline{8789} & \underline{numerical values, particularly those indicating ages or durations} \\
\textbf{11892} & \textbf{names of characters in a narrative context} \\
\underline{14512} & \underline{names and achievements of athletes, particularly in rugby} \\
15142 & names of individuals and associated figures in various contexts \\

\hline
\end{tabular}
\caption{Gemma-2-2B Repetition Features (AQ)}\label{tab:gemmafeatureaq}
\end{table*}

\begin{table*}[ht]
\centering
\begin{tabular}{|c|p{12cm}|}
\hline
\rowcolor{red!10}
\textbf{Feature ID} & \textbf{Description} \\
\hline
\multicolumn{2}{|c|}{\textbf{Layer 22}} \\
\hline
111 & references to the author and her works, focusing particularly on the name ``Taryn'' \\
1171 & the name "Shi" in various contexts \\
\textbf{3509} & \textbf{names and titles of individuals in professional contexts} \\
4999 & references to a specific name or term with the prefix ``Hy''. \\
\underline{5327} & \underline{names or mentions of a specific individual or group}\\
5521 & terms and phrases associated with epithelial growth factor receptors and related biological processes \\
11848 & phrases indicating deficiencies or absences in various contexts \\
\textbf{14137} & \textbf{phrases related to durations and periods of time} \\
14216 & phrases that indicate assertiveness and standing out or standing firm \\
\underline{15056} & \underline{certain key terms and phrases related to various subjects such as programming}, \\
& \underline{medicine, and science} \\[2pt]
\hline
\multicolumn{2}{|c|}{\textbf{Layer 24}} \\
\hline
937 & references to specific biological or medical terms and processes \\
\textbf{1119} & \textbf{numerical values or formats in mathematical or programming contexts} \\
3043 & references to specific biological or medical terms and processes \\ 
4237 & references to parenting and family dynamics \\
\underline{5333} & \underline{date ranges and time periods} \\
6707 & references to academic publications and scientific authors \\
8876 & references to significant personal events and celebrations, particularly anniversaries and milestones \\
9408 & the word ``In'' at the beginning of sentences or clauses \\
10864 & mathematical expressions and formulas related to statistical functions \\
\textbf{11892} & \textbf{names of characters in a narrative context} \\
\underline{14512} & \underline{references to modal verbs and their usage in sentences} \\
16050 & phrases indicating conditional statements or scenarios involving the subject ``we''. \\
\hline
\end{tabular}
\caption{Gemma-2-2B Repetition Features (NQ)}\label{tab:gemmafeaturenq}
\end{table*}

\begin{table*}[ht]
\centering
\begin{tabular}{|c|p{12cm}|}
\hline
\rowcolor{red!10}
\textbf{Feature ID} & \textbf{Description} \\
\hline
\multicolumn{2}{|c|}{\textbf{Layer 24}} \\
\hline
1000 & phrases indicating mathematical processes and proofs \\
\textbf{1341} & \textbf{numerical values associated with dates and times} \\
\textbf{4975} & \textbf{mathematical symbols and structures within equations} \\
7332 & titles of movies or works that include the phrase "Last" in various formats \\
8837 & references to the name "Walter" and its variations in different contexts \\
\underline{24546} & \underline{numbers associated with dates and years} \\
25636 & references to a specific individual named Russell \\
\underline{27100} & \underline{references to company names and partnerships} \\
29591 & words and phrases related to evil \\
32356 & dates and numerical values related to events \\

\hline
\multicolumn{2}{|c|}{\textbf{Layer 29}} \\
\hline
22 & phrases related to musical instruments and their cultural context \\
\textbf{4815} & \textbf{mathematical expressions and operations in formal notation} \\
\underline{11894} & \underline{character names and elements indicating romance} \\
12837 & items related to craft beer and its various qualities and attributes \\
12950 & references to proximity or closeness, both physically and metaphorically \\
\underline{13331} & \underline{elements related to mathematical concepts and programming syntax} \\
\underline{13617} & \underline{elements related to specific numerical data and coding terminology} \\
16376 & references to organizations and initiatives focused on community support and advocacy \\
\underline{21958} & \underline{references to "Game of Thrones" and related content} \\
23327 & popular television shows and their ratings \\
\underline{23499} & \underline{mathematical terminology and quantifiable data} \\
\underline{32089} & \underline{names of individuals and organizations} \\[2pt]

\hline
\end{tabular}
\caption{Llama-3.1-8B Repetition Features (EQ)}
\label{tab:llamafeatureeq}
\end{table*}

\begin{table*}[ht]
\centering
\begin{tabular}{|c|p{12cm}|}
\hline
\rowcolor{red!10}
\textbf{Feature ID} & \textbf{Description} \\
\hline
\multicolumn{2}{|c|}{\textbf{Layer 24}} \\
\hline
\textbf{1341} & \textbf{numerical values associated with dates and times} \\
1715 & mathematical symbols and expressions related to variable manipulation and equations \\
2921 & phrases related to planning and organization for events or activities \\
\textbf{4975} & \textbf{mathematical symbols and structures within equations} \\
5718 & references to durations and timing in multimedia content \\

\underline{6806} & \underline{references to specific names or entities, likely within a context of sports teams} \\
& \underline{or competitions} \\
8751 & instances of copyright-related terms and phrases \\
15453 & mathematical variables and symbols in equations \\
18162 & terms and phrases related to solar energy and sustainability initiatives \\
\underline{19305} & \underline{specific names and titles related to individuals and brands} \\

19411 & phrases related to waste management and disposal processes \\

20921 & mathematical symbols and terms related to equations and parameters \\
25861 & phrases indicating the absence or nonexistence of studies or evidence related to medical treatments and conditions \\
28578 & numerical data and formatting, particularly relating to time and monetary values \\

\hline
\multicolumn{2}{|c|}{\textbf{Layer 29}} \\
\hline
3000 & phrases related to political discussions and legislative actions \\
\textbf{4815} & \textbf{mathematical expressions and operations in formal notation} \\
7211 & instances of the pronoun ``she'' \\
8227 & the name ``John'' in various contexts \\
11475 & mentions of Wi-Fi \\
11528 & numerical data and statistics, particularly those related to measurements or scores \\
\underline{13331} & \underline{elements related to mathematical concepts and programming syntax}\\
\underline{13617} & \underline{elements related to specific numerical data and coding terminology} \\
\underline{23499} & \underline{mathematical terminology and quantifiable data} \\
26025 & terms related to search engine optimization (SEO) and digital marketing strategies \\
26421 & references to political entities, particularly countries and their governments \\

\hline
\end{tabular}
\caption{Llama-3.1-8B Repetition Features (AQ)}
\label{tab:llamafeatureaq}
\end{table*}

\begin{table*}[ht]
\centering
\begin{tabular}{|c|p{12cm}|}
\hline
\rowcolor{red!10}
\textbf{Feature ID} & \textbf{Description} \\
\hline
\multicolumn{2}{|c|}{\textbf{Layer 24}} \\
\hline

\textbf{1341} & \textbf{numerical values associated with dates and times} \\
1719 & instances of the name "Lance" and its variations\\    
2340 & terms and phrases associated with corruption and bribery \\
2443 & phrases indicating various types and classifications \\
\textbf{4975} & \textbf{mathematical symbols and structures within equations} \\
\underline{6806} & \underline{references to specific names or entities, likely within a context of sports teams} \\
& \underline{or competitions} \\
\underline{19305} & \underline{specific names and titles related to individuals and brands} \\
24418 & phrases indicating time durations and related contexts \\
\underline{24546} & \underline{numbers associated with dates and years} \\
\underline{27100} & \underline{references to company names and partnerships}\\
27990 & specific phrases or structures commonly used in formal writing or legal contexts\\

\hline
\multicolumn{2}{|c|}{\textbf{Layer 29}} \\
\hline

\textbf{4815} & \textbf{mathematical expressions and operations in formal notation} \\
6824 & names and identifiers related to individuals and family lineage \\
\underline{11894} & \underline{character names and elements indicating romance} \\
13001 & instances of the word "back" in various contexts \\
13111 & proper nouns and significant names \\
15023 & mentions of service quality and customer experiences \\
16414 & time indicators related to events \\
\underline{21958} & \underline{references to "Game of Thrones" and related content} \\
20338 & specific terminology related to medical or health-related subjects \\
28972 & expressions of excitement or emphasis \\
\underline{32089} & \underline{names of individuals and organizations} \\[2pt]

\hline
\end{tabular}
\caption{Llama-3.1-8B Repetition Features (NQ)}
\label{tab:llamafeaturenq}
\end{table*}

\section{Comparison of Repetition Features and Regular Features\label{app:D}}
To more clearly observe the presence of the repetition feature, we randomly selected a feature and compared it with one of the repetition features we identified. In Table \ref{tab:100}, when activation feature 100 was steered, the model exhibited generation behavior that matched the feature description, producing content such as 'president' related to 'political', which is a typical response after activating a regular feature. However, in Table \ref{tab:20199}, after activating feature 20199, the model's response exhibited a clear repetition phenomenon.
\begin{table*}[htbp]
\centering
\begin{tabular*}{\linewidth}{@{\extracolsep{\fill}}|c|p{12cm}|@{}}  
\hline
\rowcolor{yellow!10}
\textbf{Feature 20199} & \textbf{Names of places or people mentioned in a speaking context} \\ 
\hline
Question & Can an object have a negative distance traveled? Why or why not? \\
\hline
Normal & Can an object have a negative distance traveled? Why or why not? No, an object cannot have a negative distance traveled because distance is a scalar quantity representing the total length of the path, which is always non-negative. However, displacement, a vector quantity, can be negative as it accounts for direction and represents the straight-line distance between \\ \hline
Steered \#1 & Can an object have a negative distance traveled? Why or why not? Can an object have a negative distance traveled? Why or why not? Can an object have a negative distance travelled? Why or why not? Can an object have a negative distance travelled? Why or why not? Can an object have \\ \hline
Steered \#2 & Can an object have a negative distance traveled? Why or why not? Can an object have a negative distance traveled? Why or why not? If the object is in a positive space, it will be able to travel as long as it's visible from its first position, and the distance travelled by its \\ \hline
Steered \#3 & Can an object have a negative distance traveled? Why or why not? Can an object have a negative distance traveled? Why or why not? Can an object have a negative distance traveled? Why or why not? Can an object have a negative distance traveled? Why or why not? Can an object have \\ \hline

\hline
\textbf{1-gram repeat score} & 0.74, increased from normal by 0.51 \\ \hline
\end{tabular*}
\caption{Comparison of Activation Before and After of Feature 20199 at Layer 7\label{tab:20199}}

\end{table*}
\begin{table*}[htbp]
\centering
\begin{tabular*}{\linewidth}{@{\extracolsep{\fill}}|c|p{12cm}|@{}}  
\hline
\rowcolor{yellow!10}
\textbf{Feature 100} & \textbf{Words related to political campaigns and candidates
} \\ 
\hline
Question & When I look at myself in the mirror, I see...\\
\hline
Normal & When I look at myself in the mirror, I see a reflection of my physical appearance, which represents how others might perceive me. It shows my facial features, body posture, and expressions. However, it may not fully capture my thoughts, emotions, or inner self, which go beyond the mirror’s image. \\ \hline
Steered \#1 & When I look at myself in the mirror, I see that I'm very lucky to be a woman. I would make a great president of the United States. And if we don't have the same kind of laws, and we don't have the same  freedoms, then people will say things like \\ \hline
Steered \#2 & When I look at myself in the mirror, I see that I'm not the only person who has been able to do something that's been a lot different. The other people of this country have had some of the worst things. And  we' ve seen a president who has made some very bad choices, and \\ \hline
Steered \#3 & When I look at myself in the mirror, I see that I have a bad attitude. And if you can't be president of the United States, you' re going to have two more years. If we don't have a fair deal on immigration,  we' re going to be in for another seven years. \\ \hline

\hline
\textbf{1-gram repeat score} & 0.16, increased from normal by 0.10 \\ \hline
\end{tabular*}
\caption{Comparison of Activation Before and After of Feature 100 at Layer 7\label{tab:100}}

\end{table*}
\section{Mitigating the Repeat Curse\label{app: controlrepeatcurse}}
We demonstrate the generation effect of GPT2-small Layer 9 after deactivating 100\% of the repetition feature in Table \ref{tab:controlrepeatcurse}. In the normal (non-activated) case, when the model faces a problem requiring diversity, it falls into the repetition curse, repeatedly generating the word ``The Godfather''. However, after deactivating the repetition feature, the model is not affected by the diversity issue and does not fall into the repetition curse. In cases 2 and 3, it even shows improved diversity, listing more song information and providing more effective answers to the question.
\begin{table*}[htbp]
\centering
\begin{tabular}{|c|p{12cm}|}
\hline
Question & Could you recommend as many famous English songs as possible? \\
\hline
Normal & I recommend: \textbf{1."Losing My Mind for Grief" 2."The World Is a Beautiful Place" 3.“The Godfather”}
In case you missed it: The title "The Godfather" by Michael Jackson is probably one of the best English songs on YouTube right now! How can you possibly say "Godfather" without being wrong? You know how people think about American music when they hear it at concerts? Well, we all know that "The Godfather" is a great song.  \textbf{4.“The Godfather”5.“The Godfather”6.“The Godfather”}. \\
\hline
Unsteered\#1 & The list of the best English songs is not exhaustive, but rather it is a collection of 15 great songs that have been recorded by talented artists.
\textbf{"The Black Swan" (1955)} – The song was written by John Lennon in 1955 and has been featured on numerous TV shows and films over the years. It's one of those rare American rock albums that sounds like it's being sung to you while you're asleep on your way home from work. The main reason this one should be included in any list is because it's so well known throughout the world for its wonderful sound design, which makes this song very accessible to anyone who wants to listen at night or during long periods of time without having to worry about getting lost or forgetting what they were listening for!
\textbf{"Tina Turner" (1957)} – Tina Turner was born in New York City, where she spent her childhood growing up. \\
\hline
Unsteered\#2 & The following list is based on my own research and personal experience. I have chosen to include all of my favourite songs from the past few years, so if you're looking for a great English song that sounds like it could be included, please consider this list. If you're looking for some more classics that have been featured in the past few years, please feel free to add them here. The list also includes music from other genres such as rock (particularly rock 'n' roll), jazz (especially jazz), and pop (including pop music).
\textbf{
1."One More Time" - "Bitch In My Pocket" by Michael Jackson
2."I Can't Stop Loving You" by The Beatles
3."You Got Me" by The Beatles
4."You Ain't Got No Money Now" by The Rolling Stones
5."One More Time - I Will Always Love You" by Jack White}. \\
\hline
Unsteered\#3 & I would suggest:
\textbf{
"The Spirit of Love" by The Beatles (1962)
"Gone With the Wind" by Eminem (1975-76)
"Losing My Mind for Grief"
"The World Is a Beautiful Place"
"My Life's a Waste of Time"}. \\
\hline
\end{tabular}
\caption{Mitigate Repeat Curse (GPT2 small).}
\label{tab:controlrepeatcurse}
\end{table*}



\section{Human Evaluation\label{app:humanevaluation}}
To determine the repeat score threshold $\rho$ for the repetition feature, we manually evaluated the repetition in the generated text. If the text exhibited repetition, it was classified as ``Yes''. We randomly sampled 100 pairs of texts and then calculated the repeat score for those classified as ``Yes''. Figure \ref{tab:chooserho} displays a randomly selected portion of our evaluation process.
\begin{table*}[htbp]
\centering
\begin{tabular*}{\linewidth}{@{\extracolsep{\fill}}|c|p{12cm}|@{}}  
\hline
\rowcolor{yellow!10}
\textbf{Feature 14137} & \textbf{phrases related to durations and periods of time} \\ 
\hline
Question & How are physical quantities expressed in physics?  \\
\hline
Normal & Are leading zeros considered significant figures in a measurement?                           Here's why this is a tricky question: Leading zeros are often just placeholders. They don't represent any real quantity and don't affect the precision of the measurement. Significant figures are about representing actual \\ \hline
Steered  & How are physical quantities expressed in physics? period period period period period Period periods period periods periodsPeriod Periodperiod period Periodperiod span span span Periodnperiod period time time time time span span duration \\ \hline
Repeat Score & 0.54, increased from normal by 0.36 \\ 
\hline
Human Evaluation & Yes \\
\hline
\rowcolor{yellow!10}
\textbf{Feature 15780} & \textbf{exclamatory expressions and phrases beginning with ``Oh''.} \\
\hline
Question & If an object moves to the left, what sign would its displacement have? \\
\hline
Normal & If an object moves to the left, what sign would its displacement have? Here's a breakdown: Displacement: The overall change in position of an object. It's like the "final destination" of the movement. Distance: The total length of the path traveled.   \\
\hline
Steered & If an object moves to the left, what sign would its displacement have? dear god please help me! I am so confused about this! dear god please help me! I am so confused about this! no worries my dear, we'll get it sorted right away. yes yes yeahhhhhhh yeahhhhh  \\
\hline
Repeat Score & 0.35, increased from normal by 0.17 \\
\hline
Human Evaluation & No \\
\hline
\rowcolor{yellow!10}
\textbf{Feature 4384} & \textbf{specific terms and phrases related to research, documents, and citations} \\
\hline
Question & What are nonstandard units of measurement, and can you provide an example? \\
\hline
Normal & What are nonstandard units of measurement, and can you provide an example? Nonstandard Units of Measurement Nonstandard units are informal or arbitrary measures used to quantify certain physical quantities. They're not part of the internationally recognized system of measurement (SI) which includes standard units like meters, kilograms, and seconds. \\
\hline
Steered & What are nonstandard units of measurement, and can you provide an example? Nonstandard Units of Measurement Nonstandard units of measurement are informal or arbitrary units that are not officially recognized by a particular system of measurement. They are often used in everyday life, but they can be confusing when dealing with scientific or technical \\
\hline
Repeat Score & 0.21, increased from normal by 0.05 \\
\hline
Human Evaluation & No \\
\hline
\rowcolor{yellow!10}
\textbf{Feature 8768} & \textbf{numerical data related to time and measurements} \\ 
\hline
Question & Is displacement a vector or scalar quantity? \\
\hline
Normal & Is displacement a vector or scalar quantity? Answer: Displacement is a vector quantity. Here's why: Vector quantities have both magnitude (size) and direction. Scalar quantities only have magnitude. Displacement tells us not only how  \\
\hline
Steered &  Is displacement a vector or scalar quantity? Explanation: Vector quantities have both magnitude and direction. Scalar quantities only have magnitude. Let me know if you want to explore some specific examples! Please note that the concept of ``displacement'' \\
\hline
Repeat Score & 0.07, increased from normal by 0.01 \\ 
\hline
Human Evaluation & No \\
\hline

\end{tabular*}
\caption{A Portion of the Human Evaluation Process\label{tab:chooserho}}
\end{table*}

\section{Diversity Challenge Problem Dataset\label{app:EQ}}
To construct the question answering dataset, we employed an automated generation method augmented by manual curation. Specifically, we developed a Python script utilizing the Hugging Face Transformers library and the Meta-Llama/Llama-3.1-8B model, along with predefined prompt templates and configuration parameters, to generate an initial batch of 5,000 questions. The script first configured the environment, loaded the model, and then iteratively generated questions while performing validation and cleaning. After initial deduplication, we obtained 976 unique questions. To ensure the quality and relevance of the dataset, we conducted rigorous manual screening and ultimately selected 500 high-quality enumeration questions. Finally, these questions were saved as a JSON-formatted dataset to serve as a foundation for subsequent research.
\tcbset{
    colback=white, colframe=blue,
    width=0.95\textwidth,
    boxrule=0.8pt, arc=2mm,
    fonttitle=\bfseries,
    boxsep=5pt,
}

\begin{table*}[!ht]
    \centering
    \small
    \begin{tcolorbox}[title=Appendix: Selected List of Generated Questions (50 of 500 Questions), colback=pink!5, colframe=blue!20]
        \noindent\textbf{Question 1:} Please write down as many names as you beginning with letter 'A': Alice, Ann, Andrew?
        \vspace{0.2em}

        \noindent\textbf{Question 5:} What are the primary colors in the RGB color model?: Red, Green, Blue?
        \vspace{0.2em}

        \noindent\textbf{Question 6:} List mammals found in Africa: African Elephant, Lion, Giraffe?
        \vspace{0.2em}

        \noindent\textbf{Question 7:} List countries with monarchies: United Kingdom, Japan, Saudi Arabia?
        \vspace{0.2em}

        \noindent\textbf{Question 8:} Can you name various types of renewable energy sources?: Solar, Wind, Hydropower?
        \vspace{0.2em}

        \noindent\textbf{Question 14:} List countries in Africa starting with 'A': Algeria, Angola, Algeria?
        \vspace{0.2em}

        \noindent\textbf{Question 15:} Name flowers blooming in spring: Tulip, Daffodil, Cherry Blossom?
        \vspace{0.2em}

        \noindent\textbf{Question 18:} Write down fruits grown near the equator: Banana, Mango, Pineapple?
        \vspace{0.2em}

        \noindent\textbf{Question 20:} What are some common programming languages used in web development?: JavaScript, Python, PHP?
        \vspace{0.2em}

        \noindent\textbf{Question 22:} List programming languages used in data science: R, Python, SQL?
        \vspace{0.2em}

        \noindent\textbf{Question 25:} List vegetables that grow underground: Potato, Carrot, Onion?
        \vspace{0.2em}

        \noindent\textbf{Question 26:} Identify flowers with petals: Rose, Sunflower, Daisy?
        \vspace{0.2em}

        \noindent\textbf{Question 34:} List types of cheese: Cheddar, Brie, Gouda?
        \vspace{0.2em}

        \noindent\textbf{Question 39:} List Baroque composers: Johann Sebastian Bach, Antonio Vivaldi, George Frideric Handel?
        \vspace{0.2em}

        \noindent\textbf{Question 40:} Name colors in a rainbow: Red, Orange, Yellow?
        \vspace{0.2em}

        \noindent\textbf{Question 42:} List cities built on rivers: London, Paris, Cairo?
        \vspace{0.2em}

        \noindent\textbf{Question 47:} Identify fruits used in desserts: Apple, Strawberry, Banana?
        \vspace{0.2em}

        \noindent\textbf{Question 51:} Identify vegetables high in protein: Spinach, Broccoli, Peas?
        \vspace{0.2em}

        \noindent\textbf{Question 58:} List inventions from the Industrial Revolution: Steam Engine, Spinning Jenny, Telegraph?
        \vspace{0.2em}

        \noindent\textbf{Question 62:} List types of triangles excluding right and isosceles: Scalene Triangle, Equilateral Triangle, Obtuse Triangle?
        \vspace{0.2em}

        \noindent\textbf{Question 65:} Name extinct animal species: Dodo, Woolly Mammoth, Tasmanian Tiger?
        \vspace{0.2em}

        \noindent\textbf{Question 68:} Write down prime numbers between 1 and 15 greater than five: 7, 11, 13?
        \vspace{0.2em}

        \noindent\textbf{Question 70:} List animals with more than two legs: Centipede, Millipede, Spider?
        \vspace{0.2em}

        \noindent\textbf{Question 72:} Write down birds starting with 'C': Crow, Crane, Cardinal?
        \vspace{0.2em}

        \noindent\textbf{Question 76:} List countries with populations over 100 million: India, United States, Indonesia?
        \vspace{0.2em}

        \noindent\textbf{Question 79:} List Nobel Prize categories: Physics, Chemistry, Peace?
        \vspace{0.2em}

        \noindent\textbf{Question 80:} List mammals that lay eggs: Platypus, Echidna?
        \vspace{0.2em}

        \noindent\textbf{Question 83:} Name all the continents on Earth: Asia, Africa, Europe?
        \vspace{0.2em}

        \noindent\textbf{Question 87:} What are common types of phobias?: Arachnophobia (spiders), Claustrophobia (confined spaces), Acrophobia (heights)?
        \vspace{0.2em}

        \noindent\textbf{Question 88:} Name elements used in solar panels: Silicon, Silver, Aluminum?
        \vspace{0.2em}

        \noindent\textbf{Question 89:} List as many Olympic sports as possible: Athletics, Swimming, Gymnastics?
        \vspace{0.2em}

        \noindent\textbf{Question 92:} Name elements in the noble gas group: Helium, Neon, Argon?
        \vspace{0.2em}

        \noindent\textbf{Question 96:} List countries with significant coral reefs: Australia (Great Barrier Reef), Philippines, Indonesia?
        \vspace{0.2em}

        \noindent\textbf{Question 98:} Please list as many countries in South America as you can: Brazil, Argentina, Colombia?
        \vspace{0.2em}

        \noindent\textbf{Question 100:} Write down languages spoken in India: Hindi, Bengali, Tamil?
        \vspace{0.2em}

        \noindent\textbf{Question 101:} Name fruits used in alcoholic beverages: Grape (wine), Apple (cider), Juniper (gin)?
        \vspace{0.2em}

        \noindent\textbf{Question 103:} Name plants that are not grasses: Oak Tree, Rose, Cactus?
        \vspace{0.2em}

        \noindent\textbf{Question 104:} Name sports in Olympic competitions: Basketball, Soccer, Tennis?
        \vspace{0.2em}

        \noindent\textbf{Question 108:} List cryptocurrency names: Bitcoin, Ethereum, Litecoin?
        \vspace{0.2em}

        \noindent\textbf{Question 111:} Identify flowers used in perfumes: Rose, Jasmine, Lavender?
        \vspace{0.2em}

        \noindent\textbf{Question 112:} Identify types of cloud formations: Cumulus, Stratus, Cirrus?
        \vspace{0.2em}

        \noindent\textbf{Question 113:} List countries where Spanish is the official language: Spain, Mexico, Argentina?
        \vspace{0.2em}

        \noindent\textbf{Question 114:} Write down types of renewable energy: Solar, Wind, Geothermal?
        \vspace{0.2em}

        \noindent\textbf{Question 122:} List countries beginning with 'G': Germany, Ghana, Greece?
        \vspace{0.2em}

        \noindent\textbf{Question 123:} What are popular social media platforms?: Facebook, Instagram, TikTok?
        \vspace{0.2em}

        \noindent\textbf{Question 131:} Identify fruits containing vitamin C: Orange, Kiwi, Guava?
        \vspace{0.2em}

        \noindent\textbf{Question 132:} Write down US states bordering an ocean: California, Florida, Maine?
        \vspace{0.2em}

        \noindent\textbf{Question 135:} Write down major deserts in Africa: Sahara, Kalahari, Namib?
        \vspace{0.2em}

        \noindent\textbf{Question 136:} Write down types of clouds: Cumulus, Stratus, Cirrus?
        \vspace{0.2em}

        \noindent\textbf{Question 141:} Name citrus fruits: Lemon, Grapefruit, Lime?
        \vspace{0.2em}

        \noindent\textbf{Question 143:} List cities with historic fortifications: Carcassonne, York, Dubrovnik?
        \vspace{0.2em}
    \end{tcolorbox}
    \caption{Selected List of Generated Questions (50 of 500 Questions)}
    \label{tab:Appendix-Selected-Questions-Boxed-v6}
\end{table*}

\section{Utility After Mitigation\label{app:New}} 
We conduct two case studies to further investigate utility after mitigation. The feature description will be shown in Table \ref{tab:new}.

\noindent {\bf Deactivating Features in Universal Question} We identified ten repetitive features associated with names in Llama-3.1-8B’s 24 layers, located at indices: [12656, 8575, 7468, 15812, 10614, 916, 1866, 10781, 16247, 640]. These features were deactivated for analysis.

We then selected three questions—one from each of our three datasets—likely to be heavily influenced by name-related features. The results in Table \ref{tab:case1} showed that while some repetition was reduced, the overall answer quality remained unaffected.

\begin{table*}[htbp]
\centering
\begin{tabular}{|l|p{12cm}|}
\hline
\rowcolor{red!10}
\multicolumn{2}{|c|}{\textbf{Case Study 1: Deactivating Features in Universal Question}} \\
\hline
\multicolumn{2}{|c|}{\textbf{EQ}} \\
\hline
Question & Please write down as many names as you beginning with letter 'A': Alice, Ann, Andrew \\
\hline
\multicolumn{2}{|c|}{\textbf{After deactivation}} \\
\hline
Output & Please write down as many names as you beginning with letter 'A': Alice, Ann, Andrew, Anna. Can you think of any more? Please send them to me if you can. \\
\hline
Repeat Score & 0.133333 \\
\hline
Information Entropy & 0.940527 \\
\hline
\multicolumn{2}{|c|}{\textbf{Before deactivation}} \\
\hline
Output & Please write down as many names as you beginning with letter 'A': Arthur, Adam, Alan, Alex, Anthony, Andrew, Arthur, Adam, Alan, Alex, Anthony, Andrew, Arthur, Adam, Alan, Alex, Anthony, Andrew, Arthur, Adam \\
\hline
Repeat Score & 0.619048 \\
\hline
Information Entropy & 0.655519 \\
\hline
\multicolumn{2}{|c|}{\textbf{NQ}} \\
\hline
Question & Do you know where the Golden Gate Bridge is \\
\hline
\multicolumn{2}{|c|}{\textbf{After deactivation}} \\
\hline
Output & Do you know where the Golden Gate Bridge is located? -- It’s in San Francisco, California. The Golden Gate Bridge is a suspension bridge that connects San Francisco to Marin County. It was built in 1937 and has become one of the \\
\hline
Repeat Score & 0.166667 \\
\hline
Information Entropy & 0.938184 \\
\hline
\multicolumn{2}{|c|}{\textbf{Before deactivation}} \\
\hline
Output & Do you know where the Golden Gate Bridge is located? Do you know what the Golden Gate Bridge looks like? If not, then this is for you! This bridge was built in 1937 and has been a major tourist attraction ever since. \\
\hline
Repeat Score & 0.190476 \\
\hline
Information Entropy & 0.929353 \\
\hline
\multicolumn{2}{|c|}{\textbf{AQ}} \\
\hline
Question & Do you know Lagrange's mean value theorem \\
\hline
\multicolumn{2}{|c|}{\textbf{After deactivation}} \\
\hline
Output & Do you know Lagrange's mean value theorem? Let's see the following 1. If we have a differentiable function f(x) on the closed interval [a, b], then there exists at least one point c in (a, \\
\hline
Repeat Score & 0.028571 \\
\hline
Information Entropy & 0.988859 \\
\hline
\multicolumn{2}{|c|}{\textbf{Before deactivation}} \\
\hline
Output & Do you know Lagrange's mean value theorem? It is a very useful theorem in calculus. The proof is also quite easy. But it took me more than 10 minutes to find out the version of it, which is suitable for my needs \\
\hline
Repeat Score & 0.047619 \\
\hline
Information Entropy & 0.979005 \\
\hline
\end{tabular}
\caption{Case Study 1: Deactivating Features in Universal Question}
\label{tab:case1}
\end{table*}

\noindent {\bf Deactivating Features in Specific Question}
We attempted to identify proper noun in the 24th layer and deactivating the features related to the word and asking questions associated with it, studying whether our suppression affected the generation outcome. We selected the word "Rome" and "Vatican" and obtained the following related features: [21185, 25206, 2152, 26865, 1108, 17963], where 21185, 25206, 17963 are repetition features. The results in Table \ref{tab:case2} showed that the deactivation did not impact the model's generation of knowledge.

\begin{table*}[htbp]
\centering
\begin{tabular}{|l|p{12cm}|}
\hline
\rowcolor{red!10}
\multicolumn{2}{|c|}{\textbf{Case Study 2: Deactivating Features in Specific Question}} \\
\hline
Question & Where is the Vatican located in Rome\\
\hline
\multicolumn{2}{|c|}{\textbf{After deactivation}} \\
\hline
Output & The Vatican is located in the heart of Rome, on the west bank of the Tiber River. It is a sovereign city-state within Italy and covers an area of just over \\
\hline
Repeat Score & 0.161290 \\
\hline
Information Entropy & 0.925056 \\
\hline
\multicolumn{2}{|c|}{\textbf{Before deactivation}} \\
\hline
Output & Where is the Vatican located in Rome? The Vatican City (the Holy See) is not a province of Italy, but an independent state. It is located on the west bank of the Tiber River, within the city limits of Rome. \\
\hline
Repeat Score &  0.224999 \\
\hline
Information Entropy & 0.898956 \\
\hline
\end{tabular}
\caption{Case Study 2: Deactivating Features in Specific Question}
\label{tab:case2}
\end{table*}
\begin{table*}[ht]
\centering
\begin{tabular}{|c|p{12cm}|}
\hline
\rowcolor{red!10}
\textbf{Feature ID} & \textbf{Description} \\
\hline
\multicolumn{2}{|c|}{\textbf{Case Study 1}} \\
\hline
12656 & numerical data and specific names related to statistical or regulatory information \\
8575 & variations of the word ``Le'' as in names or titles \\
7468 & words associated with company names and mergers\\
15812 & mentions of individuals named John F. Kennedy and variations of that name \\
10614 & mentions of family-related names and relationships \\
916 & instances of the name ``Johnny.'' \\
1866 & information related to full names and dates \\
10781 & references to the name ``Jason'' and its variations \\
16247 & corporate names and their corresponding locations or identifiers \\
640 & proper names of authors or researchers \\
\hline
\multicolumn{2}{|c|}{\textbf{Case Study 2}} \\
\hline
21185 & references to specific locations and landmarks in Rome \\
25206 & locations and geographical references \\
2152 & geographical locations and addresses \\
507 & geographic locations, particularly cities in Asia \\
21185 & references to specific locations and landmarks in Rome \\
25206 & locations and geographical references \\
2152 & geographical locations and addresses \\
26865 & references to the Vatican and associated terms \\
1108 & references to Saint Peter and related locations in the Vatican \\
17963 & terms related to legal and canonical authority within the context of the Vatican \\

\hline
\end{tabular}
\caption{Feature Discription of Two Case Studies for Utility Evaluation}\label{tab:new}
\end{table*}

\noindent {\bf Discussion} Farrell E \citep{farrell2024applying} argues that ``negative scaling of feature activations is necessary and zero ablating features is ineffective.'' However, in our experiments, we found that setting the activation value to a small positive activation value (0.01) achieves good mitigation effects while preserving the model's generative capabilities. Unlike negative scaling—which induces knowledge unlearning—our approach of using minimal positive activations successfully mitigates repetition without compromising the model's knowledge retention. 
Thus, I suggest these two deactivation value choices serve distinct purposes:
\begin{itemize}
    \item Negative scaling is suitable for removing hazardous knowledge (e.g., harmful memorized content).
    \item Minimal positive activation is preferable for behavioral mitigation (e.g., reducing repetition) while retaining model functionality.
\end{itemize}

\end{document}